\pdfoutput=1

\documentclass[11pt]{article}

\usepackage[preprint]{acl}

\usepackage{times}
\usepackage{latexsym}

\usepackage[T1]{fontenc}

\usepackage[utf8]{inputenc}

\usepackage{microtype}

\usepackage{inconsolata}

\usepackage{graphicx}
\usepackage{tcolorbox}

\usepackage{makecell}
\usepackage{subcaption} 

\usepackage{mfirstuc}

\usepackage{tabularray}
\UseTblrLibrary{booktabs} 
\UseTblrLibrary{siunitx}  

\usepackage{amsmath,amsfonts,bm}

\def\eqref#1{equation~\ref{#1}}

\def\1{\bm{1}}

\def\vx{{\bm{x}}}
\def\vy{{\bm{y}}}

\DeclareMathAlphabet{\mathsfit}{\encodingdefault}{\sfdefault}{m}{sl}
\SetMathAlphabet{\mathsfit}{bold}{\encodingdefault}{\sfdefault}{bx}{n}

\DeclareMathOperator*{\argmin}{arg\,min}

\def\draftmodel{M_q}
\def\targetmodel{M_p}
\def\wsamplingpolicy{S_W}

\def\windowlength{h}
\def\whatcost{ensembling costs}

\def\tabedmt{MT*}
\def\tabedmtcp{MTCP*}
\usepackage{comment}

\usepackage{xcolor}
\usepackage{soul}

\definecolor{customyellow}{HTML}{FFFF01}

\newcommand{\hlc}[2][yellow]{{%
    \colorlet{foo}{#1}%
    \sethlcolor{foo}\hl{#2}}%
}

\usepackage{amsmath,amsfonts,bm}

\usepackage{booktabs}
\usepackage{multirow}
\usepackage{adjustbox}
\usepackage{rotating}
\usepackage{tikz}
\usepackage{subcaption}
\usetikzlibrary{positioning, shapes, arrows}

\colorlet{accept}{blue}
\usepackage{siunitx}

\usepackage{algorithm}
\usepackage{algorithmicx, algpseudocode}

\usepackage{pgfplots}
\pgfplotsset{compat=1.18}

\newcommand{\wonjun}[1]{\textcolor{green}{Wonjun: #1}}
\newcommand{\minjae}[1]{\textcolor{orange}{Minjae: #1}}

\newcommand{\ba}[1]{\textcolor{red}{Byeongkeun: #1}}
\newcommand{\kl}[1]{\textcolor{red}{Kangwook: #1}}
\usetikzlibrary{shapes,arrows,positioning,calc,patterns}

\usepackage{wrapfig}

\usepackage{colortbl}

\colorlet{mygold}{yellow!30}
\colorlet{mysilver}{black!10}
\colorlet{mybronze}{white}

\usepackage{enumitem}

\usepackage{subcaption}

\def\ours{TABED}
\def\oursnotation{\ours{}\textsuperscript{MT}}

\def\oursadaens{TABED} 
\def\oursfull{Test-time Adaptive Batched Ensemble Drafting}
\def\ourspeedup{1.74}
\def\ourspeedupmt{5}
\def\ourimprovement{5}
\def\numw{n}

\def\singleabbv{Sgl.}
\def\ensembleabbv{Ens.}

\def\targetmodel{M_p} 
\def\draftmodel{M_q} 
\def\chunklength{\gamma}

\def\random{Random}
\def\static{Static}

\def\mtrandom{\random{}\textsuperscript{MT}} 
\def\mt{\static{}\textsuperscript{MT}} 

\def\mtcprandom{\random{}\textsuperscript{MTCP}} 
\def\mtcp{\static{}\textsuperscript{MTCP}} 
\def\oursmtcpnotation{\ours{}\textsuperscript{MTCP}}

\def\treewidth{d}

\def\vision{vision encoder}
\def\prefill{prefilling}
\def\decode{decoding}

\def\standard{benchmark}
\def\ood{OOD}
\def\nlpdataset{NLP Datasets}

\def\draft{drafting}
\def\verify{verification}

\usetikzlibrary{shapes.symbols}

\usepackage{titletoc}
\usepackage{titlecaps}
\usepackage{amsthm}
\newtheorem{remark}{Remark} 

\usepackage{mathtools}

\newcommand{\best}{\textbf}
\newcommand{\secondbest}{\underline}
\newcommand{\citenumb}[1]{[\citenum{#1}]}

\colorlet{mymax}{gray!15}

\DeclareRobustCommand\onedot{\futurelet\@let@token\@onedot}
\def\@onedot{\ifx\@let@token.\else.\null\fi\xspace}
\def\eg{\emph{e.g.}} 
\def\ie{\emph{i.e}\onedot}

\usepackage[colorinlistoftodos, linecolor=red, backgroundcolor=yellow!20, bordercolor=red]{todonotes}
\setuptodonotes{size=\footnotesize}

\newtcolorbox{highlight}[1][]{
    colback=yellow!15,
    colframe=gray!30,
    boxrule=1pt,
    arc=2pt,
    leftrule=2pt,
    rightrule=2pt,
    toprule=0.5pt,
    bottomrule=0.5pt,
    #1
}
\usepackage[capitalize]{cleveref}

\NewDocumentCommand{\mj}
{ mO{} }{\textcolor{blue}{\textsuperscript{\textit{M}}\textsf{\textbf{\small[#1]}}}}

\title{\ours{}: Test-Time Adaptive Ensemble Drafting\\for Robust Speculative Decoding in LVLMs}

\newcommand*\samethanks[1][\value{footnote}]{\footnotemark[#1]}

\author{
Minjae Lee$^{1}$\thanks{Equal contribution.}
\hspace{0.3em}
Wonjun Kang$^{1}$\samethanks{}\hspace{0.3em}
Byeongkeun Ahn$^{1}$
\hspace{0.3em}
Christian Classen$^{2}$
\hspace{0.3em}
\textbf{Kevin Galim}$^{1}$
\hspace{0.3em}
\\
\textbf{Seunghyuk Oh}$^{1}$
\hspace{0.3em}
\textbf{Minghao Yan}$^{2}$
\hspace{0.3em}
\textbf{Hyung Il Koo}$^{1}$
\hspace{0.3em}
\textbf{Kangwook Lee}$^{2,3}$
 \\
$^{1}$ FuriosaAI \hspace{0.3em} $^{2}$ UW-Madison \hspace{0.3em} $^{3}$ KRAFTON \\
{\tt\small \{minjae.lee, kangwj1995\}@furiosa.ai, kangwook.lee@wisc.edu}
}

\begin{document}
\maketitle
\begin{abstract}
Speculative decoding (SD) has proven effective for accelerating LLM inference by quickly generating draft tokens and verifying them in parallel.
However, SD remains largely unexplored for Large Vision-Language Models (LVLMs), which extend LLMs to process both image and text prompts.
To address this gap, we benchmark existing inference methods with small draft models on 11 datasets across diverse input scenarios and observe scenario-specific performance fluctuations.
Motivated by these findings, we propose \emph{\oursfull{} (\ours{})}, which dynamically ensembles multiple drafts obtained via batch inference by leveraging deviations from past ground truths available in the SD setting.
The dynamic ensemble method achieves an average robust walltime speedup of \ourspeedup{}× over autoregressive decoding and a \ourimprovement{}\% improvement over single drafting methods, while remaining training-free and keeping \whatcost{} negligible through parameter sharing.
With its plug-and-play compatibility, we further enhance \ours{} by integrating advanced verification and alternative drafting methods.
Code and custom-trained models are available at \url{https://github.com/furiosa-ai/TABED}.
\end{abstract}

\begin{figure*}[t]
    \centering
    \includegraphics[width=\textwidth]{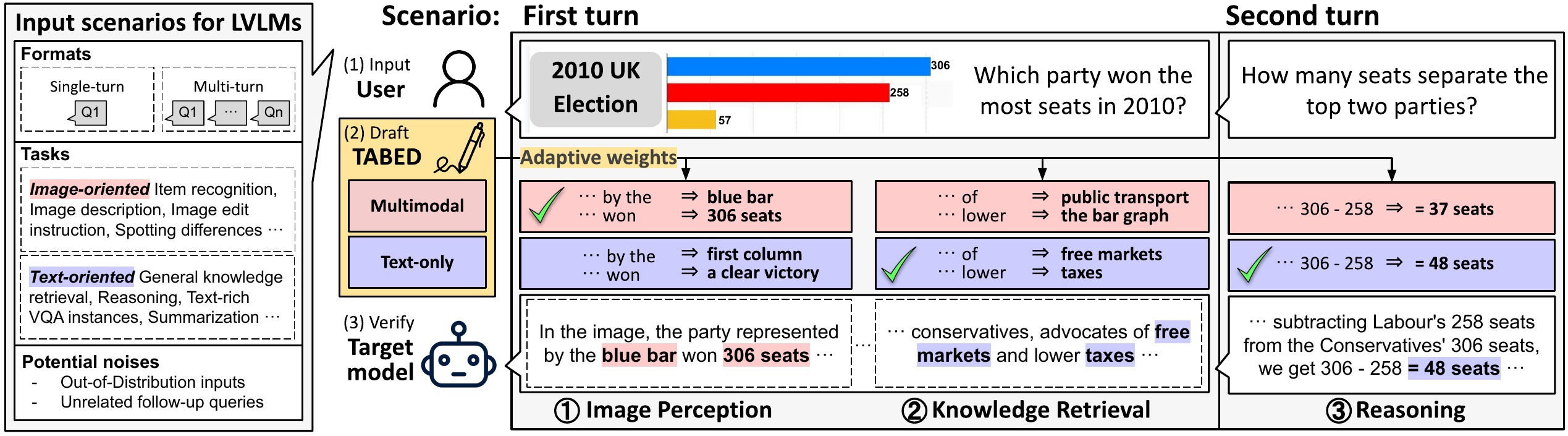}
\caption{
\textbf{Overview of \ours{}.}
LVLMs must handle diverse input scenarios involving various combinations of turn-taking, tasks, and potential noise. To effectively accelerate target LVLMs with SD, different drafting methods are required for intra- and inter-response cases.
For example, in the first turn, the drafting method needs to identify the visual context referenced by the user based on the input image and then perform text-based knowledge retrieval (\raisebox{.5pt}{\textcircled{\raisebox{-.9pt}{1}}} → \raisebox{.5pt}{\textcircled{\raisebox{-.9pt}{2}}}). In the second turn, it performs mathematical reasoning using the accumulated text context (\raisebox{.5pt}{\textcircled{\raisebox{-.9pt}{3}}}).
Across such varying scenarios, existing single drafting methods with small models—whether multimodal or text-only—exhibit fluctuating performance (\cref{sec:study}).
\textbf{\oursadaens{}} addresses this by dynamically ensembling multiple drafts using past ground truths in SD, achieving robust speedups across diverse scenarios (\cref{sec:ensemble}). It can be further enhanced by integrating advanced verification and drafting methods (\cref{sec:further}). See \cref{fig:tta_ibed} for details.
}
\vspace{-0.5em}

\label{fig:scenarios}
\end{figure*}
\section{Introduction}


Multimodal Large Language Models (MLLMs) \cite{wu2023multimodallargelanguagemodels, zhang2024mmllmsrecentadvancesmultimodal,jin2024efficientmultimodallargelanguage, song2023bridgegapmodalitiescomprehensive} are an advanced class of LLMs~\citep{brown2020gpt3,ouyang2022training,touvron2023llama,guo2025deepseek} designed to process multiple modalities, such as images, audio, and video, alongside text.
In particular, Large Vision Language Models (LVLMs)~\cite{chen2024dressinstructinglargevisionlanguage}, also known as Large Multimodal Models~\cite{li2024llava,song2023bridgegapmodalitiescomprehensive}, specialize in handling prompts comprised of text and images. 
These models have attracted significant attention due to their unique applications, including multimodal chatbots, visual question answering (VQA), and augmented reality (AR)~\citep{openai2023gpt4,claude,geminiteam2023gemini}. 

As LVLMs are increasingly deployed, reducing their inference time has become a critical challenge.
In addition to the standard LLM autoregressive decoding process, LVLMs must (1) pre-process each image in the input into several hundred {image tokens}~\citep{radford2021clipvit,liu2023llava,liu2024improved} and (2) process both text and image tokens, resulting in considerably higher inference time.
Recently, methods like token pruning, layer skipping, and key-value cache compression have been proposed to accelerate LVLM inference \citep{shang2024llava,chen2024image,lin2024boosting,liu2024efficient,wan2024look}. 
While effective, these approximation techniques cannot preserve the original LVLM's output distribution. 
Moreover, they primarily reduce prompt processing time (i.e., the \prefill{} stage) and have less impact on response generation time (i.e., the \decode{} stage).

In contrast, Speculative Decoding (SD) is an acceleration technique for LLM inference that fully preserves the output distribution~\citep{leviathan2023fast, chen2023accelerating}.
SD {speculates} a specified number of draft tokens and then uses the original target model to {verify} these tokens in parallel. 
For LLM inference, various methods have been proposed to enhance different components of SD~\citep{xia2024unlockingefficiencylargelanguage,hu2025speculative}, such as aligning a lightweight draft model more closely with the target model~\citep{zhou2024distillspec,kang2025vispec,hu2025adaspec}, or constructing token trees to verify multiple drafts simultaneously using a target model augmented with additional components~\citep{cai2024medusa,li2024eagle,li2025eagle}.
These approaches yield substantial performance gains but often require additional training and reduce compatibility with other acceleration methods, thereby limiting deployment accessibility.
Unlike in LLMs, SD for LVLMs has been far less explored.
\citet{gagrani2024speculative} is the first to successfully accelerate LVLM inference via SD using a text-only drafting method that relies solely on the LLM component of an LVLM and text input tokens.
While this approach is training-free and its findings are both intriguing and counterintuitive, it is evaluated only on a few single-turn VQA datasets, limiting insight into when the method performs well without image inputs.
Another notable work, \citet{kang2025vispec}, makes the draft model vision-aware by compressing image tokens and injecting a global visual context.
However, it requires a dedicated data generation using the target model, followed by training the draft model.
This raises the question:
\emph{``Can we develop a robust, vision-aware, and training-free drafting method that delivers consistent acceleration across diverse real-world input scenarios?''}
To answer this, we first benchmark existing inference methods for drafting at scale on seven benchmark datasets and two out-of-distribution (OOD) datasets using multi-turn instruction interactions.
We find that none of the existing inference methods with a small draft model—whether multimodal or text-only~\citep{gagrani2024speculative}—effectively handle diverse input scenarios, making the selection of a single drafting method in advance a nontrivial challenge.
To address this, we present \oursfull{} (\ours{}), which (1) obtains multiple drafts simultaneously via batch inference and (2) dynamically ensembles them based on deviations from the ground truth provided by the target model. The method is entirely training-free and fully compatible with enhanced SD approaches in a plug-and-play manner.
As a result, \oursadaens{} consistently achieves the best or second-best performance across all scenario combinations, with an average robust expected walltime speedup of \ourspeedup{}× over standard decoding, without any additional tuning. It delivers over \ourimprovement{}\% improvement compared to existing single drafting methods, which exhibit scenario-specific performance fluctuations.
We further enhance \ours{} by integrating advanced verification and alternative drafting methods, leveraging its plug-and-play compatibility.
All components operate in a fully training-free manner with negligible \whatcost{}, as demonstrated empirically.
To further support LVLM SD research and ensure reproducibility, we open-source our code and draft LVLMs.

In summary, main contributions of our work are:
\begin{itemize}[leftmargin=*]

\item We benchmark existing inference methods for drafting and find that their performance fluctuates across diverse LVLM input scenarios (Sec.~\ref{sec:study}).
\item We propose a dynamic ensemble drafting method, \ours{}, which achieves superior and robust performance across diverse input scenarios (Sec.~\ref{sec:ensemble}).
\item We further enhance \ours{} by integrating advanced verification and alternative drafting methods via its plug-and-play compatibility (Sec.~\ref{sec:further}).

\end{itemize}

\section{Related Work}
\label{sec:related}
\subsection{Large Vision Language Models}
\paragraph{LVLMs}
Frontier proprietary LVLMs~\citep{openai2023gpt4, claude, geminiteam2023gemini} exhibit state-of-the-art performance across multiple modalities beyond just text. Meanwhile, open-source models such as the LLaVA series \citep{liu2023llava, liu2024improved, li2024llava, li2024llava-onevision}, Qwen-VL series~\citep{wang2024qwen2,bai2025qwen2}, and LLaMA 3.2~\citep{dubey2024llama} are also rapidly advancing. 

While various methods exist for embedding image inputs~\citep{yin2024surveymultimodallargelanguage, jin2024efficientmultimodallargelanguage}, one of the most prominent approaches, LLaVA, employs an off-the-shelf vision encoder~\citep{radford2021clipvit,zhai2023sigmoidlosslanguageimage} and a trainable projector to convert each image into several hundred visual context tokens for an LLM.

\paragraph{Approximate Inference}
To address the inefficiency of handling visual tokens from images,
several approaches have been proposed based on a common finding: only a sparse subset of the hundreds of visual tokens is important, enabling reduced computational cost with minimal information loss. \citet{shang2024llava, chen2024image, lin2024boosting} dynamically prune significant visual tokens based on attention sparsity. Further focusing on reducing redundant key-value caches, \citet{liu2024efficient,wan2024look} retain key-value vectors by merging or discarding less critical caches during inference. However, from a latency perspective, these approaches primarily benefit the \prefill{} stage while providing limited advantages for the \decode{} stage.

\subsection{Speculative Decoding}

\paragraph{SD for LLMs}

SD accelerates LLM inference using a small draft model while preserving the target model's output distribution \citep{leviathan2023fast, chen2023accelerating}. 
To improve the \draft{} phase, various efforts have been made, including generating multiple draft candidates~\citep{miao2023specinfer,sun2024spectrfastspeculativedecoding,yang2024multi,li2025eagle}, and fine-tuning the draft model with knowledge distillation~\citep{zhou2024distillspec,hu2025adaspec}. 
Some studies employ speculative approaches to address exceptionally long sequence lengths that significantly degrade decoding efficiency~\citep{sun2024triforce,chen2024magicdec,galim2025draft}. 
Separately, other work investigates newer architectures for SD, including State Space Models (SSMs)~\citep{wang2024mamba,choi2025mamba}.

\paragraph{SD for LVLMs}
\citet{gagrani2024speculative} introduces text-only drafting, which uses only the language model component of an LVLM with text input.
Without any additional training of the draft model (i.e., training-free), it achieves performance comparable to multimodal drafting.
However, their benchmark results and analyses are limited, and they do not explore how to effectively leverage image information for improved drafting. Moreover, whether multiple drafting methods can be effectively combined remains unclear.
\citet{kang2025vispec} present a notable work to make the draft model more vision-aware. Their method involves compressing image tokens and injecting a global visual context into the draft model. However, this approach requires a dedicated data generation phase using the target model, followed by a training process for the draft model.
\citet{jang2024lantern} and \citet{teng2024accelerating} propose SD methods for text-to-image generation, which is distinct from LVLM approaches.

\subsection{Test-time Adaptation}

To enhance model robustness against distribution shifts during test time, various generalization and adaptation techniques have been proposed~\citep{liang2025comprehensive}. In situations where data arriving during test time lacks ground truth labels, these techniques update models through explicit gradient steps by formulating loss using pseudo labeling~\citep{hardt2024testtimetrainingnearestneighbors, wang2022continual}, self-supervision~\citep{sun2020testtimetrainingselfsupervisiongeneralization, krause18a2018dynamicevaluation}, or normalization statistics~\citep{schneider2020improving, wang2021tent}, thereby causing inevitable delays during test time. 
In contrast, SD allows the use of ground-truth information obtained through verification for test-time adaptation. In this regard, \citet{xia2024swift} enables test-time adaptation for SD but incurs high adaptation cost, limiting its applicability to infrequent domain shifts.

\section{Preliminaries: Speculative Decoding}
\label{sec:prelim}

\label{sec:spec_decode_latency}

\paragraph{Algorithm}
Following~\citep{leviathan2023fast, zhou2024distillspec}, let $M_p$ be the target model whose inference we aim to accelerate, and let $M_q$ be the draft model for the same task. For a given prefix $x$, decoded sequence $y_{<t}$, block length $\gamma$, and in-block index $n = 0, \ldots, \gamma-1$, the following steps are repeated until either an \texttt{<EOS>} token is accepted or the maximum sequence length is reached:
\begin{enumerate}[leftmargin=*]
\item The \textit{\titlecap{\draft{} phase}}, where $M_q$ sequentially generates $\gamma$ draft tokens $y_{{t+n}} \sim q(\cdot | x, y_{<{t+n}})$.
\item The \textit{\titlecap{\verify{} phase}}, where $M_p$ reviews these draft tokens in parallel, comparing them to $p(y_{{t+n}} | x, y_{<{t+n}})$.
\item For sampling, each token $y_{t+n}$ is sequentially accepted with probability $\min\left(1, \frac{p(y_{{t+n}} | x, y_{<{t+n}})}{q(y_{{t+n}} | x, y_{<{t+n}})}\right)$. If any token is rejected before the end of the block, subsequent tokens are discarded, and the first rejected token is resampled from the adjusted distribution $\text{norm}(\max(0, p(y) - q(y)))$.\footnote{When the prefix \((x, y_{<t})\) is clear from the context, we use \( p(y) \) and \( q(y) \) to denote \( p(y_t | x, y_{<t}) \) and \( q(y_t|x, y_{<t}) \), respectively.}
\end{enumerate}

\paragraph{Block Efficiency and Walltime Speedup}
The \textit{block efficiency}, denoted as $\tau_{p,q}(\gamma)$, is defined as the expected number of accepted tokens per block.  
The expected \textit{walltime speedup} quantifies the ratio between the per-token inference time of autoregressive decoding with $\targetmodel$ and that of the speculative decoding method.  
For a given batch size $B$ and sequence length $S$, let $T_p(B,S)$ and $T_q(B,S)$ denote the time required for $M_p$ and $M_q$ to decode a single token, respectively, and let $T_p(B,S,\gamma)$ denote the time required for $M_p$ to verify $\gamma$ tokens in parallel.  
For brevity, we omit $(B,S)$ and use the simplified notations $T_p$, $T_q$, and $T_p(\gamma)$.  
When $B$ and $S$ are small, the speculative decoding time per block for producing $\tau_{p,q}(\gamma)$ valid tokens can be approximated as $T_{\mathrm{SD}} = \gamma \cdot T_q + T_p(\gamma) \approx \gamma \cdot T_q + T_p$~\citep{chen2024magicdec}.
Under this approximation, the expected walltime speedup is given by
\begin{equation}
\vspace{-0.25em}
\textit{Speedup} = \frac{T_p}{T_{\mathrm{SD}} / \tau_{p,q}(\gamma)} 
\approx \frac{\tau_{p,q}(\gamma)}{\gamma \cdot \frac{T_q}{T_p} + 1}.
\label{eq:speedup}
\vspace{-0.25em}
\end{equation}

We first empirically measure the draft-to-target latency ratio ${T_q}/{T_p}$, and then use the computed speedup in~\cref{eq:speedup} to evaluate the runtime efficiency of speculative decoding. 
See~\cref{app:remarks} for empirical measurement of the latency ratio ${T_q}/{T_p}$ and analysis of its effects on latency and throughput.
\begin{table*}[t]
\centering
\sisetup{table-auto-round}
\renewcommand{\arraystretch}{1.0}
\caption{
Block efficiency results for drafting methods are shown, including existing single-drafting approaches—multimodal (M) and text-only (T)—and our proposed method, {\textbf{\oursnotation{}}}, which dynamically ensembles M and T through test-time adaptation.
Across diverse combinations of turn-taking settings, datasets (both benchmark and OOD), image inclusion, and contextual relatedness, \oursnotation{} consistently achieves either the \best{best} or \secondbest{second-best} performance, while M and T vary depending on the scenario. Even when ranked second, its performance remains close to the best, unlike the larger gaps observed in other methods.
}
\label{tab:acl_main}
\resizebox{1.0\textwidth}{!}{%
\begin{tabular}{ll|cccccccc||cc}
\toprule
\multicolumn{2}{c|}{Drafting} &
\multicolumn{8}{c||}{\titlecap{\standard{} datasets} (First Turn)} &
\multicolumn{2}{c}{\titlecap{\ood{} datasets}} \\
\cmidrule(lr){1-2}\cmidrule(lr){3-10}\cmidrule(lr){11-12}
Type & Method & LLaVA-W & DocVQA & POPE & MMVet & IEdit & MB & Spot & Avg. & PSV & VIST \\
\midrule
\multirow{2}{*}{Single} & M & \best{2.28} & \secondbest{2.15} & \best{2.56} & \best{2.21} & 2.19 & 1.96 & \secondbest{2.34} & \secondbest{2.24} & 1.19 & 1.16 \\
& T~\citenumb{gagrani2024speculative} & 2.19 & 2.08 & 2.31 & 2.16 & \secondbest{2.23} & \secondbest{2.34} & 2.27 & 2.23 & \best{2.05} & \best{2.05} \\
\cmidrule(lr){1-12} 
\multirow{1}{*}{Ensemble} & \textbf{\oursnotation{}} & \secondbest{2.26} & \best{2.16} & \secondbest{2.52} & \best{2.21} & \secondbest{2.29} & \best{2.39} & \best{2.36} & \best{2.31} & \secondbest{2.02} & \secondbest{2.04} \\
\midrule
\multicolumn{2}{c|}{Drafting} &
\multicolumn{8}{c||}{\titlecap{\standard{} datasets} (Second Turn)} &
\multicolumn{2}{c}{\titlecap{\nlpdataset{}}} \\
\cmidrule(lr){1-2}\cmidrule(lr){3-10}\cmidrule(lr){11-12}
Type & Method & LLaVA-W & DocVQA & POPE & MMVet & IEdit & MB & Spot & Avg. & NQ & GSM8K \\
\midrule
\multirow{2}{*}{Single} & M & 2.10 & 1.96 & 2.78 & 2.18 & 1.61 & 1.53 & 1.83 & 2.00 & 1.98 & 2.25 \\
& T~\citenumb{gagrani2024speculative} & \best{2.32} & \best{2.23} & \secondbest{2.91} & \best{2.56} & \best{1.87} & \best{2.01} & \best{2.08} & \best{2.28} & \best{2.03} & \best{2.30} \\
\cmidrule(lr){1-12} 
\multirow{1}{*}{Ensemble} & \textbf{\oursnotation{}} & \secondbest{2.29} & \best{2.23} & \best{2.93} & \best{2.56} & \secondbest{1.85} & \secondbest{1.99} & \secondbest{2.05} & \secondbest{2.27} & \best{2.03} & \secondbest{2.29} \\
\bottomrule
\end{tabular}
}
\end{table*}

\section{Benchmarking SD for LVLMs}
\label{sec:study}
In this section, we systematically benchmark speculative decoding for LVLMs by evaluating existing drafting methods, both multimodal and text-only, across a variety of input scenarios.


\subsection{Experiment Settings}


\paragraph{Target and Draft Models}
We employ 7B and 13B models from the LLaVA-1.5~\citep{liu2024improved} and LLaVA-NeXT~\citep{liu2024llavanext} series, both of which are widely adopted public LVLMs, as our target models.
We then use three distinct variants of the draft model, considering different model sizes and training strategies to ensure comprehensive benchmarking.
Due to the lack of sufficiently small LVLMs that ensure a low \textit{draft-to-target latency ratio} (represented by $\frac{T_q}{T_p}$ in \cref{eq:speedup}), we select the small public LLaMA models of 68M and 160M \citep{miao2023specinfer} as base models for training.
We then train the draft model using different training methodologies, specifically utilizing the procedures for LLaVA-1.5, as well as LLaVA-OV \citep{li2024llava-onevision}, which are specialized in multi-image processing. 
We further verify that the draft model captures multimodality through representative multimodal tasks (see \cref{sec:app_train_detail} for training details and \cref{sec:app_draft_evaluation} for evaluation results).

To summarize, we benchmarked draft models by size (68M, 160M) and type (LLaVA-1.5, LLaVA-OV), and target models by size (7B, 13B) and type (LLaVA-1.5, LLaVA-NeXT).
For brevity, we primarily report results with the LLaVA-1.5 68M–7B pair, unless otherwise noted.

\paragraph{Input Scenarios}
Systems using SD must deliver consistent performance gains across diverse real-world scenarios, especially due to their application with LVLMs in practical settings.
To examine this requirement, we initially curated seven benchmark datasets due to the lack of pre-existing benchmarks specifically for LVLM SD. These datasets include tasks involving both single-image~\citep{liu2023llava,mathew2021docvqa,li2023evaluating,yu2023mm} and multi-image scenarios~\citep{tan2019expressing,zhang2024magicbrush,jhamtani2018learning}. 
To further challenge the system's ability to handle unexpected and varied queries while maintaining consistent performance, we added two additional datasets featuring five images per query~\citep{li2019storygan,huang2016visual}, serving as notable Out-of-Distribution (OOD) cases.
Moreover, we extend the evaluation from single-turn to multi-turn scenarios using benchmark datasets that include various types of follow-up queries, including those dependent on prior responses (e.g., follow-up requests with images from the same dataset or text-only tasks from multi-turn benchmarks for LVLMs~\citep{liu2024convbenchmultiturnconversationevaluation}) and distinct text-only reasoning tasks~\citep{kwiatkowski2019natural,cobbe2021training}. 

\paragraph{Drafting Methods: Multimodal and Text-only}
The multimodal drafting method follows the standard LVLM process, receiving both images and text as input. In contrast, the text-only drafting method, first explored in~\citet{gagrani2024speculative}, uses only textual data, adhering to the standard LLM process. 
We primarily report results with $\gamma=5$ and greedy decoding with a maximum of 128 new tokens, unless otherwise noted.

\subsection{Experimental Results}
\cref{tab:acl_main} presents the block efficiency results for multimodal (M) and text-only (T) drafting methods.
Across all scenarios, neither drafting method consistently outperformed the other. Notably, the multimodal drafting method demonstrated higher block efficiency than the text-only approach in single-turn scenarios across most \standard{} datasets. However, the text-only drafting performed comparably or slightly sub-optimally overall in single-turn situations, often surpassing the multimodal drafting in subsequent turns--particularly in tasks dependent on prior responses and independent reasoning tasks, and significantly in OOD cases. This trend continued regardless of the model size or multi-image awareness capabilities. 
Existing drafting methods, both multimodal and text-only, show scenario-specific performance fluctuations with draft models small enough for SD, with neither method consistently outperforming the other across various input scenarios for LVLM inference. 
Before execution, it is challenging to determine in advance which method is superior, and even if known, resolving inconsistencies with a single drafting method is difficult.
See~\cref{app:generalization,app:ensemble_mt} for full results for various $\draftmodel{}$ and $\targetmodel{}$ sizes and types, and different $\chunklength{}$, are provided.

\begin{highlight}
    \emph{No one-size-fits-all among existing single drafting methods for LVLM input scenarios.}
\end{highlight}

\section{\oursfull{}}

\label{sec:ensemble}
For LVLM SD, selecting a single drafting method in advance for a small draft model is challenging. In this context, ensemble learning offers a promising solution, as it is often used to reduce both bias and variance in predictions ~\citep{dietterich2000ensemble,ganaie2022ensemble}, particularly for models with limited capacity ~\citep{zhou2012ensemble}. 

\begin{figure}[!t]
    \centering
    \includegraphics[width=\linewidth]{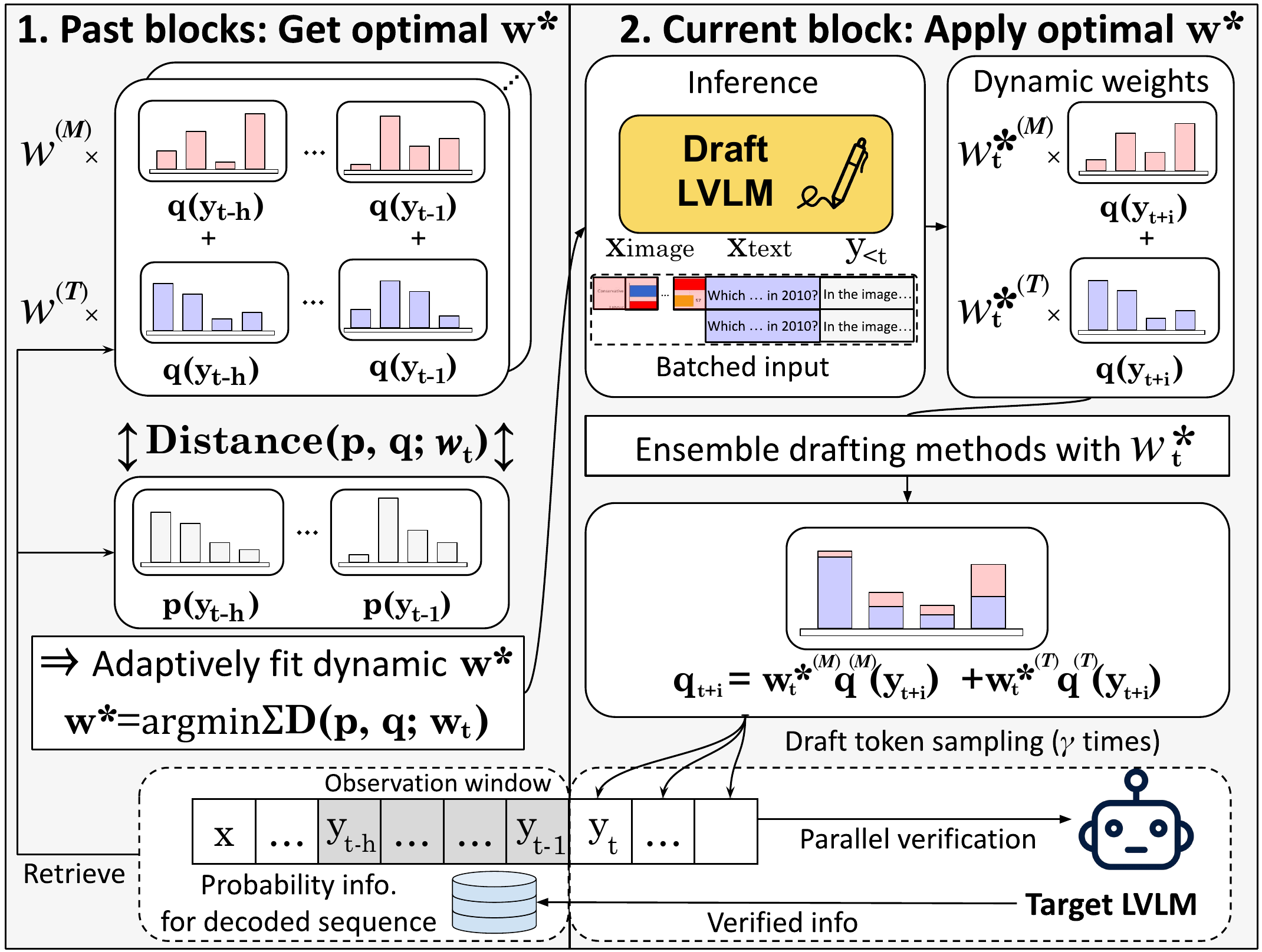}
\caption{
\oursadaens{} predicts the optimal weight, $w^*$, based on the deviation (i.e., distance) of past drafting blocks from the ground truth obtained through the verification in SD.
It then dynamically ensembles multiple drafts obtained via batch inference. 
See Algorithm~\ref{ensemblealgo} for further details.
}
\vspace{-0.5em}
\label{fig:tta_ibed}
\end{figure}

In this section, we introduce \oursfull{} (\ours{}), a method that applies ensemble learning with dynamic weight adaptation to each of the drafts obtained via batch inference.
It is fully training-free, highly extensible, and delivers robust performance across diverse input scenarios. \cref{fig:tta_ibed} provides an overview of \ours{}.

\subsection{Proposed Method}
Unlike typical test-time scenarios where incoming data lacks ground truth labels~\citep{wang2021tent,wang2022continual}, SD allows access to both hard labels (i.e., $y_{<t}$) and soft labels (i.e.,  $p(y_{t'})$) for $t' < t$ after verifying all steps prior to $t$ using the target model $M_p$. 
When drafting is performed in step $t$, this information can be leveraged to dynamically adjust ensemble weights $w_t$, controlling each drafting method's influence based on its past performance. 
This approach can enhance ensemble learning by effectively combining probabilities $[q^{(1)}_{t}, \ldots, q^{(m)}_{t}]$ from $m$ drafting methods, ensuring the resulting ensembled distribution $q(\cdot|x,y_{<t}; w_t)$ closely aligns with the target distribution $p(\cdot|x, y_{<t})$. 
For instance, the system can adjust the weight assigned to the multimodal drafting method by recognizing varying needs for visual context for a specific sample $x$ and timestep $t$.


\paragraph{Ensemble learning via Batched Inference}

To obtain multiple predictions from draft model $\draftmodel{}$, we utilize batched inference tailored for LVLMs. 
At each decoding timestep \(t\) to \(t + \gamma - 1\), all $m$ drafting methods share the parameters of $\draftmodel$, and their distributions are ensembled to sample the next token (Algorithm~\ref{ensemblealgo}). We employ a weighted averaging ensemble method, sampling a token from the ensembled distribution to continue drafting. 
This process incurs no additional costs during training (i.e., training-free) and keeps \whatcost{} negligible, unlike typical ensemble learning that requires multiple extra models and computational overhead for deployment.

\paragraph{Test-time Adaptive Ensemble Weights}

To effectively combine drafting methods based on varying needs, we explore test-time adaptation to dynamically weight each method.
Specifically, at the beginning of each new drafting block, we sample a list of ensemble weights $W_t = [w_t^{1}, \ldots, w_t^{n}]$  using the weight sampling policy $S_W(t)$ (\cref{ensemblealgo}). 
For each $w_t^j\in \mathbb{R}^{m}$, we compute ensembled draft probabilities $q_{t'}^j = q(\cdot|x, y_{<t'}; w_t^j)$ over the past time step window $t' \in [t-\windowlength{}, t-1]$. 
We then select the optimal $w^*_t$ from $W_t$ to be used at timestep $t$ by utilizing hard labels $y_{<t}$ (by maximizing the number of token matches sampled from each $q_{t'}^j$), or soft labels $ p(y_{t'})$ (by minimizing the accumulated error $e_t^j$ over previous steps $t'$).

$$e_t^j = \sum_{t'} D_{\text{KL}}\big(p(\cdot \mid x, y_{<t'}) \parallel q(\cdot \mid x, y_{<t'};w_t^j)\big)$$

where \(D_{\text{KL}}\) is the KL divergence between \(p\) and \(q_{t'}^j\) at each of the previous steps $t'$. 
This weight \(w_t^{*}\) is used throughout the current drafting block of \(\gamma\) tokens (i.e., from timestep \(t\) to \(t + \gamma - 1\)). 
To avoid redundant computation during drafting, we cache and reuse the distributions $q_{t'}^j$ from the $\draftmodel$ and the labels $p(y_{t'})$ from the $\targetmodel$, computed over the past time-step window $t' \in [t-\windowlength{},\, t-1]$.

Drafting methods with higher weights indicate closer alignment to the target model. 
In our experiments, we also explored the effects of employing various weight sampling policies $\wsamplingpolicy{}$, varying the window size $\windowlength{}$ for past time steps, and using Total Variation Distance (TVD) instead of KL divergence.

\begin{algorithm}[!t]
\caption{\ours{}}
\textbf{Parameter}: $\draftmodel{}$, Prefix $X = [x^{(1)}, ... , x^{(m)}]$ for $m$ drafting methods, Weight sampling policy $\wsamplingpolicy{}$, local window length $\windowlength{}$\\
\Comment{\textsc{W-Avg} stands for Weighted Average} \\
\Comment{\textsc{\textit{D}} stands for Distance} \\



\textbf{Input}: Batch $b_{<t}\coloneqq[(x^{(1)},y_{<t}), ... , (x^{(m)},y_{<t})]$ \\
\textbf{Output}: $\gamma$ draft tokens $y_{t}, ... , y_{t+\gamma-1}$ and ensembled probabilities $q_t$, ..., $q_{t+\gamma-1}$
\begin{algorithmic}[1]
\Procedure{\ours{}}{$b_{<t};\gamma$, $\wsamplingpolicy{}$}
    \State $W_t = [w_t^{1}, \ldots, w_t^{n}] \sim \wsamplingpolicy{}(t)$
    \State $w_t^* \gets \argmin_{j}\sum_{t'\in[t-\windowlength{}, t-1]}\textsc{\textit{D}}(p_{t'}, q_{t'}^j)$
    \For{$i \gets 0$ \textbf{to} $\gamma-1$}
        \State $[q^{(1)}_{t+i}, ... , q^{(m)}_{t+i}] \gets M_q (b_{<t+i})$ 
        \State $q_{t+i} \gets \textsc{W-Avg}($$[q^{(1)}_{t+i}, ... , q^{(m)}_{t+i}]; w_t^*$) 
        \State $y_{t+i} \gets \textsc{Sample}(q_{t+i})$ 
    \EndFor
    \State \textbf{return} $[y_t, ..., y_{t+\gamma-1}], [q_t, ..., q_{t+\gamma-1}]$
\EndProcedure
\end{algorithmic}
\label{ensemblealgo}
\end{algorithm}

\subsection{Experimental Results}

\cref{tab:acl_main} presents the block efficiency results for \oursadaens{}, which is denoted by \oursnotation{}.
\oursnotation{} consistently achieved either the best or second-best performance across all input scenarios.
These scenarios include turn-taking, image inclusion, prior context relevance, and both benchmark and OOD datasets.
Even when \oursnotation{} ranked second, its performance remained close to the best, unlike the larger performance gaps observed for other methods.
Without incurring additional costs during training or inference, this resulted in and a \ourimprovement{}\% improvement compared to existing single drafting methods M and T, when averaged across all benchmark datasets and turns. This leads to an average robust walltime speedup of \ourspeedup{}×, computed using~\cref{eq:speedup}, compared to standard decoding.
These consistent gains highlight \oursadaens's ability to effectively prioritize stronger drafting methods by dynamically assigning optimal weights, even when individual methods exhibit scenario-specific performance fluctuations.
For full results on varying the parameters of \ours{} (sampling policy $\wsamplingpolicy{}$, window size $\windowlength{}$, and distance $D$), as well as a detailed derivation of the walltime speedup with empirical analysis on latency and throughput, see~\cref{app:ensemble_mt,app:remarks}, respectively.
\section{Extensions and Analyses of \ours{}}
\label{sec:further}
\begin{table}[!t]
\centering
\sisetup{table-auto-round}
\renewcommand{\arraystretch}{1.0}
\small
\caption{
Averaged block efficiency results for single (Sgl.) and ensemble (Ens.) drafting methods integrated with token-tree verification using tree width $\treewidth$.
\textbf{\oursnotation{}} consistently achieves the \best{best} or \secondbest{second-best} performance, while M and T vary across scenarios.
See~\cref{app:token-tree} for full results.
}
\label{tab:width-avg}
\vspace{-0.5em}
\begin{tabular}{c|cl|cc}
\toprule
\multicolumn{1}{c|}{\multirow{1}{*}{\shortstack{Verification}}} &
\multicolumn{2}{c}{\multirow{1}{*}{\shortstack{Drafting}}} &
\multicolumn{2}{|c}{Block Efficiency} \\
\cmidrule(lr){1-1}\cmidrule(r){2-3}\cmidrule(l){4-5}
\multicolumn{1}{c|}{Tree width} & \multicolumn{1}{c}{Type} & \multicolumn{1}{c}{Method} & \multicolumn{1}{|c}{Benchmark} & OOD \\
\midrule

\multirow{3}{*}{\makecell{$\treewidth=2$}}& \multirow{2}{*}{\singleabbv{}}& M & \secondbest{2.89} & 1.30 \\
& & T~\citenumb{gagrani2024speculative} & 2.85 & \best{2.72} \\
\cmidrule(r){2-3}\cmidrule(l){4-5}
& \multirow{1}{*}{\ensembleabbv{}}& \textbf{\oursnotation{}} & \best{2.99} & \secondbest{2.64} \\
\midrule
\multirow{3}{*}{$\treewidth=3$} & \multirow{2}{*}{\singleabbv{}}& M & \secondbest{3.28} & 1.38 \\
& & T~\citenumb{gagrani2024speculative}& 3.24 & \best{3.19} \\
\cmidrule(r){2-3}\cmidrule(l){4-5}
& \multirow{1}{*}{\ensembleabbv{}}& \textbf{\oursnotation{}} & \best{3.39} & \secondbest{3.09} \\
\bottomrule
\end{tabular}
\vspace{-1em}
\end{table}

Since \ours{} is plug-and-play and training-free, it remains fully compatible with and can benefit from enhanced SD~\citep{xia2023speculative,hu2025speculative}.
In this section, we extend \ours{} by incorporating an enhanced verification method and integrating newly explored drafting candidates.
We then provide in-depth analyses and ablations to qualitatively validate the contribution of the proposed components.


\subsection{Further Extensions of \ours{}}

\paragraph{\titlecap{Verification: Integrating Token Tree}}

We integrate the well-known token-tree verification 
method~\citep{miao2023specinfer,yang2024multi,cai2024medusa,li2025eagle} into \ours{}, where all draft sequences from the draft model $\draftmodel$ are represented as a token tree and verified in parallel by the target model $\targetmodel$. Specifically, we employ tree widths $\treewidth = 2$ and $3$ and report the results.
\cref{tab:width-avg} presents block efficiency results averaged across datasets in each category for single drafting methods (M and T) and the dynamic ensemble method (\ours{}) combined with token-tree verification. \oursnotation{} consistently achieves the best or second-best performance, while M and T vary across scenarios—reproducing the trend observed in~\cref{tab:acl_main} with an even larger block-efficiency gap. Full results are provided in~\cref{app:token-tree}.

\paragraph{\titlecap{Drafting: Exploring Additional Candidates}}

\label{sec:extensible}

\begin{table}[t]
\centering
\sisetup{table-auto-round,mode=text}
\renewcommand{\arraystretch}{1.0}
\small
\caption{
Averaged block efficiency results for single drafting methods (M, T, caption-based (C), and pooled-multimodal (P)) and the dynamic ensemble method \oursmtcpnotation{}.
\textbf{\oursmtcpnotation{}} consistently achieves the \best{best} performance in both the first and second turns across all benchmark and OOD datasets, followed by C, which consistently achieves the \secondbest{second-best} performance.
See~\cref{app:generalization} for full results.
}
\vspace{-0.5em}
\label{tab:acl_main2}
{%
\begin{tabular}{ll|cc|c}
\toprule
\multicolumn{2}{c|}{\multirow{1}{*}{\shortstack{Drafting}}} &

\multicolumn{2}{c}{\titlecap{\standard{}}} & 
\multicolumn{1}{|c}{\multirow{2}{*}{\makecell{\titlecap{\ood{}}}}} \\
\cmidrule(lr){1-2}\cmidrule(lr){3-4}
\multicolumn{1}{c}{Type} & \multicolumn{1}{c|}{Method} & First & Second &     \\
\midrule
\multirow{4}{*}{Single} & M & 2.24 & 2.00 & 1.18 \\
& T~\citenumb{gagrani2024speculative} & 2.23 & 2.28 & 2.05 \\
& C & \secondbest{2.29} & \secondbest{2.30} & \secondbest{2.09} \\
& P & 2.23 & 2.25 & 2.08 \\
\midrule
Ensemble & \textbf{\oursmtcpnotation{}} & \best{2.32} & \best{2.32} & \best{2.13} \\
\bottomrule
\end{tabular}
}
\vspace{-1em}
\end{table}

Since handling sparsely important image tokens is a unique challenge for LVLMs—absent in typical LLMs~\citep{shang2024llava,chen2024image}—we explore effective strategies to address this issue and extend \ours{} accordingly.

\emph{\titlecap{Pooled Multimodal Drafting (P)}}
To condense sparsely important image tokens while preserving their 2D spatial structure, we apply average pooling during inference immediately before the projector maps them into the text embedding space.
Specifically, we use a $2\times2$ pooling kernel, reducing the number of visual tokens from 576 to 144 in our default configuration.

\emph{\titlecap{Caption Drafting (C)}}
To convert sparsely important image tokens into textual descriptions, we use a lightweight captioning model, Florence-2~\citep{xiao2024florence}, executed once in parallel with the \prefill{} stage of the target model $\targetmodel$.
This makes its overhead negligible and amortized over the entire decoding process (\cref{fig:timespan}).
For details on replacing image tokens in caption drafting and an analysis of captioning-model latency, see~\cref{app:caption,app:system_prompts}.

\cref{tab:acl_main2} presents block efficiency results averaged across datasets in each category for the newly explored single drafting methods (P and C) and the dynamic ensemble method \oursmtcpnotation{}.
\oursmtcpnotation{} consistently achieves the highest speedup across all benchmark and OOD datasets, outperforming all single drafting methods (M, T, C, and P) in both the first and second turns.
This seamless integration improves generalization~\citep{zhou2012ensemble} and better aligns the ensemble distribution with the target distribution, without additional training cost, while keeping \whatcost{} negligible through parameter sharing.
While P and C introduce image awareness into the drafting process, C in particular retains T’s robustness and performs second-best on both standard and OOD datasets.
See~\cref{app:generalization,app:remarks} for full results and analysis on latency and throughput of \oursmtcpnotation{}.

\subsection{Analysis and Ablations of \ours{}}
\begin{table}[!t]
\centering
\sisetup{table-auto-round}
\renewcommand{\arraystretch}{1.0}
\small
\caption{
Averaged block efficiency of ensemble drafting methods with different weighting strategies (\mtrandom{}, \mt{}, \oursnotation{}) applied to M and T. \textbf{\oursnotation{}} consistently achieves performance that is equal to or the \best{best}. The \secondbest{second-best} result is underlined.
}
\label{tab:ablation-mt}
\vspace{-0.5em}
\begin{tabular}{cc|cc|c}
\toprule
\multicolumn{2}{c|}{\multirow{1}{*}{\shortstack{Drafting}}} &
\multicolumn{2}{c|}{\titlecap{\standard{}}} &
\multicolumn{1}{c}{\multirow{2}{*}{\shortstack{\titlecap{\ood{}}}}} \\
\cmidrule(lr){1-2}\cmidrule(lr){3-4}
Type & Weights & First & Second & \\
\midrule
\multirow{3}{*}{Ensemble} & \mtrandom{} & 2.29 & 2.17 & 1.83 \\
 & \mt{} & \best{2.31} & \secondbest{2.25} & \secondbest{1.93} \\
 & \textbf{\oursnotation{}} & \best{2.31} & \best{2.27} & \best{2.03} \\
\bottomrule
\end{tabular}
\vspace{-1em}
\end{table}

\paragraph{Effectiveness of Dynamic Weights}
To assess the role of dynamic ensemble weights, we conduct an ablation study in which they are replaced with non-negative, sum-to-one random weights (\mtrandom{}) or static equal weights (\mt{}), used to combine multiple drafts generated via batched inference. 
\cref{tab:ablation-mt} presents block efficiency results averaged across datasets in each category for these ablations, and \oursnotation{} achieves performance equal to or better than both \mtrandom{} and \mt{}.
In particular, \oursnotation{} clearly outperforms \mt{} in OOD scenarios by effectively down-weighting the influence of underperforming drafting methods.
See~\cref{app:generalization} for full results across different $\draftmodel$ and $\targetmodel$ sizes.

\paragraph{{Adaptation Behavior of Dynamic Weights}}
\begin{figure}[t]
    \centering
    \includegraphics[width=\linewidth]{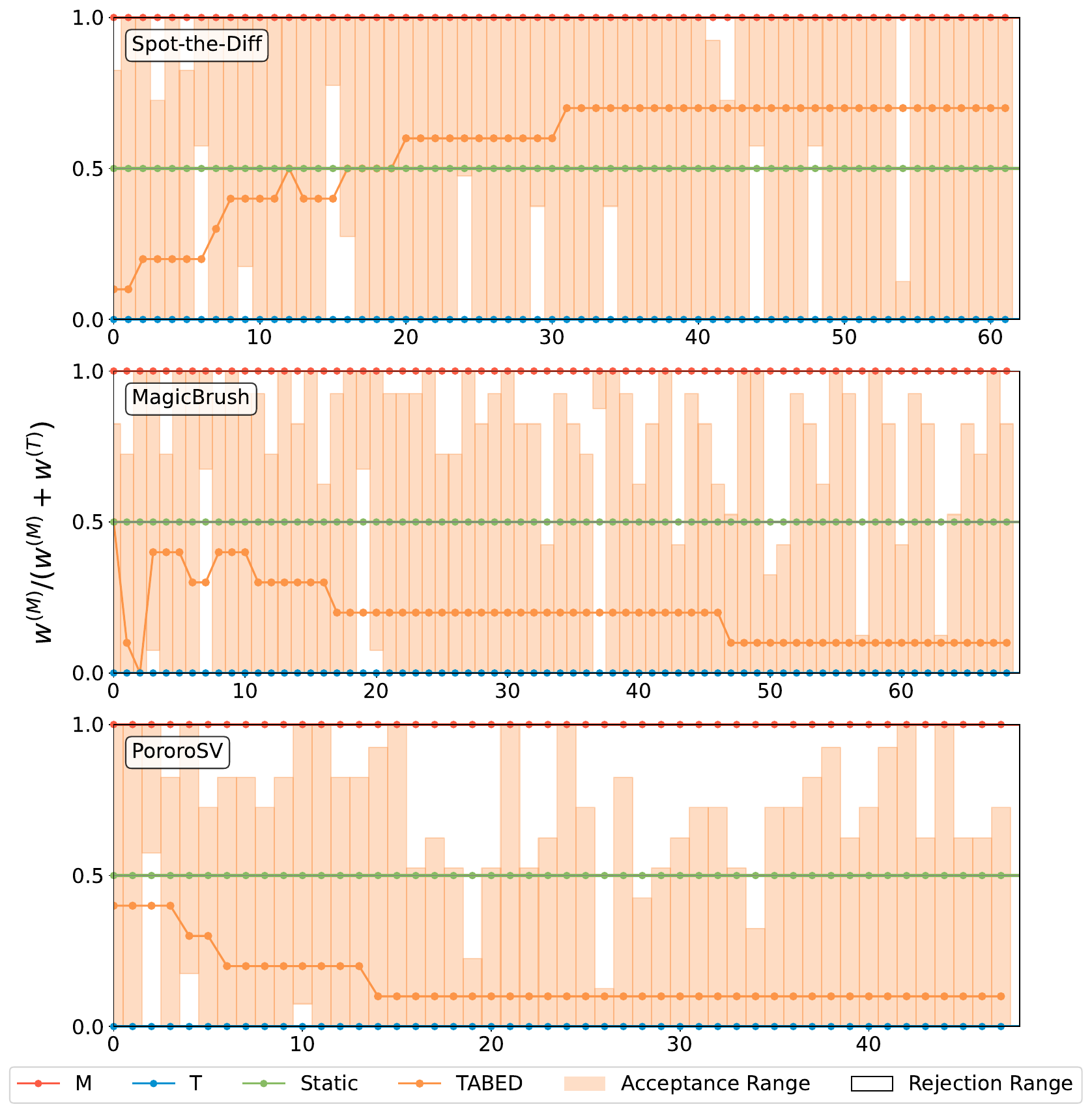}
\caption{
Qualitative samples visualizing dynamic ensemble weights across datasets. The x-axis and y-axis represent decoding steps and the proportion of $w^{(M)}$ relative to $w^{(M)} + w^{(T)}$, respectively. 
The acceptance range (shaded) comprises weights under which $\targetmodel{}$ accepts the drafted tokens, while the rejection range (unshaded) comprises weights under which it rejects them.
See \cref{app:quali_w} for more examples.
}
\vspace{-1em}
\label{fig:quali_adaptive_weights}
\end{figure}


We provide an in-depth view of how the adaptive weights $w^*$ predicted by \oursadaens{} evolve during decoding. \cref{fig:quali_adaptive_weights} visualizes the weights used by single drafting methods (M and T at $y=1$ and $0$) and ensemble methods (\mt{} and \oursnotation{} at $y=0.5$ and the orange line) across decoding steps. The weights $w^*$ predicted by {\oursnotation{}} effectively stay within the shaded acceptance range while avoiding the unshaded rejection range, unlike single drafting methods (M and T) or the static ensemble \mt{}. This adaptive behavior enables \oursnotation{} to discriminate effectively among multiple drafting methods, explaining the large performance gap observed under OOD settings in~\cref{tab:ablation-mt}. More examples are provided in \cref{app:quali_w}.

\section{Conclusion}
We benchmark existing inference methods for LVLM drafting and observe scenario-specific performance fluctuations. To address this, we propose \ours{}, that dynamically ensembles multiple drafts obtained via batch inference by leveraging deviations of the drafts from past ground truths. \oursadaens{} consistently achieves the best or second-best performance across all scenarios, with an average robust expected walltime speedup of \ourspeedup{}× and over \ourimprovement{}\% improvement over single drafting methods, all without additional tuning. \ours{} can be further enhanced through advanced verification or alternative drafting methods via its plug-and-play compatibility and remains fully training-free with negligible \whatcost{}.

\section{Limitations and Future Work}
While we focus on a plug-and-play, training-free inference method for open-source LVLMs, \ours{} can also leverage trained draft models~\citep{zhou2024distillspec,yang2024multi,hu2025adaspec}, as it is fully compatible with such approaches. 
Moreover, it would be valuable to evaluate a broader range of LVLM families (e.g., Qwen-VL and LLaMA 3.2), beyond LLaVA, as well as to explore scaled-up draft models $\draftmodel$ with billions of parameters rather than limiting experiments to small-capacity draft LVLMs.
A more theoretically grounded approach to modeling the relationship between the past time-step window and the dynamic ensemble weights could further improve the method.
We further expect \ours{} to generalize to other MLLMs (e.g., those handling audio~\cite{fu2024vita}).
Finally, recent work has explored speculative decoding~\citep{gao2025self} for diffusion-based LLMs~\citep{llada,ye2025dream} by leveraging their parallel decoding properties~\citep{kang2025parallelbench}. Extending our approach to diffusion MLLMs~\citep{llada-v} is a promising direction for future work.
\section*{Acknowledgement}
Kangwook Lee is supported by NSF Award DMS-2023239, NSF CAREER Award CCF-2339978, Amazon Research Award, and a grant from FuriosaAI.
We thank all members of the FuriosaAI team for their support, with special thanks to Hanjoon Kim and June Paik for their vision and commitment to research.
We also thank the Lee Lab at UW–Madison, led by Kangwook Lee, for their collaboration.
Finally, we thank Seongsu Bae, Eunbyeol Cho, Gyubok Lee, and Sungjin Park at KAIST EDLAB, led by Edward Choi, for their valuable comments and inspiration.

\section*{Reproducibility Statement}
The code used for all experiments in this paper is publicly available. To support reproducibility, both the code and checkpoints can be accessed at the following GitHub repository: \url{https://github.com/furiosa-ai/TABED}.

\bibliography{reference.bib}

\clearpage
\setcounter{remark}{0}
\appendix
\section*{\Large Appendix}

We structure the appendix as follows.

In~\cref{app:generalization}, we present additional generalization results.
In~\cref{app:ensemble_mt}, we provide implementation details.
In~\cref{app:quali_w}, we show qualitative examples of dynamic ensemble weights.
In~\cref{app:token-tree}, we report additional results with token-tree verification.
In~\cref{app:caption}, we describe the caption-drafting setup.
In~\cref{app:samples}, we evaluate the target model and provide sample outputs.
In~\cref{app:remarks}, we provide supporting experiments for remarks in the main text.
In~\cref{app:training}, we document draft-model training details.
In~\cref{app:system_prompts}, we list dataset- and method-specific prompt formats.
In~\cref{app:benchmark}, we describe benchmark and OOD datasets.
In~\cref{app:inference-time}, we present additional runtime analyses.

\section{Comprehensive Benchmarking Across Draft and Target Models}
\label{app:generalization}
In this section, we present the full results corresponding to the main experiments shown in~\cref{tab:acl_main,tab:acl_main2} for all drafting methods considered in this paper, including both single (multimodal and text-only) and ensemble approaches (\tabedmt{} and \tabedmtcp{}), across various sizes and types of draft and target models. Specifically, we benchmark draft models (LLaVA-1.5 68M, LLaVA-1.5 160M, LLaVA-OV 68M) and target models (LLaVA-1.5 7B, LLaVA-1.5 13B, LLaVA-NeXT 7B, LLaVA-NeXT 13B). As shown in~\cref{tab:rebt1,tab:rebt2,tab:rebt3,tab:rebt4,tab:rebt5}, the findings from~\cref{sec:study,sec:ensemble,sec:extensible} hold across different combinations of draft and target models.

\section{Implementation Details of \ours{}}
\label{app:ensemble_mt}

In this section, we examine two key factors influencing the dynamic ensembling approach: the observation window length $h$ used for computing adaptive weights (\cref{sec:app_obs_window}), the weight sampling policy $S_w$ that determines how ensemble weights are selected in our experiments (\cref{sec:app_sw}), and the draft block length $\gamma$ in the sequential draft phase (\cref{sec:app_gamma}). Finally, we complement our empirical ablations with a theoretically grounded variant, {TABED-AdaBoost}, a multi-class AdaBoost-inspired weighting scheme (\cref{sec:app_adaboost}).

\subsection{Observation Window Length $h$}
\label{sec:app_obs_window}
The responsiveness of the dynamic weights predicted by \ours{} can be controlled by the observation window length $h$. In~\cref{ensemblealgo}, we introduce an observation window of length $h$ for the drafting method to compute dynamic weights based on information within this window. By varying $h$, we further examine how this approach addresses the inherent nature of autoregressive decoding, which relies solely on past context for next-token prediction. For both benchmark and OOD datasets, we explore the effect of varying $h$ by setting it to 1, 4, 16, and to include all previous tokens. As shown in \cref{tab:app_window}, there is no clear winner among the different window lengths $h$. Therefore, by default, using all previous information (i.e., ALL) before the current timestep is a reasonable choice.

\begin{table*}[!h]
\centering
\sisetup{table-auto-round}
\renewcommand{\arraystretch}{0.9}
\caption{Experimental results with different observation window length $h$, which constrain the number of previous timesteps used to compute adaptive weights. $h=1$ indicates that the weights are based only on the ground truth from the previous decoding step.}
\label{tab:app_window}
\resizebox{1.0\textwidth}{!}
{%
\begin{tabular}{lrl|ccccccc||cc}
\toprule
\multicolumn{3}{c|}{Draft Model} & \multicolumn{7}{c||}{\titlecap{\standard{} datasets} } & \multicolumn{2}{c}{\titlecap{\ood{} datasets}} \\
\cmidrule(lr){1-3}\cmidrule(lr){4-10}\cmidrule(lr){11-12}
Type & Size & $h$ & LLaVA-W & DocVQA & POPE & MMVet & IEdit & MB & Spot & PSV & VIST \\
\midrule
\multirow{4}{*}{\shortstack[l]{LLaVA 1.5}} & \multirow{4}{*}{68M} & 1 & 2.26 & 2.15 & 2.54 & 2.21 & 2.30 & 2.36 & 2.35 & 2.00 & 2.02 \\
 &  & 4 & 2.25 & 2.16 & 2.54 & 2.20 & 2.29 & 2.40 & 2.36 & 2.05 & 2.05 \\
  &  & 16& 2.26 & 2.17 & 2.53 & 2.21 & 2.31 & 2.42 & 2.36 & 2.04 & 2.05\\
  &  & ALL& 2.26 & 2.16 & 2.52 & 2.21 & 2.29 & 2.39 & 2.36 & 2.02 & 2.04\\
\bottomrule
\end{tabular}
}


\end{table*}


\subsection{Weight Sampling Policy $S_w$}
\label{sec:app_sw}
There are numerous variants of \ours{}, obtained by varying the weight sampling policy $S_w$ in \cref{ensemblealgo}. Here, we describe the representative policy used in our experiments throughout the paper. For $\wsamplingpolicy{}$ in \tabedmt{}, we sample $W_t = [w_t^{j}]$, where $w_t^j = [1 - \frac{j}{n}, \frac{j}{n}] \in \mathbb{R}^{m}$ and $n = 10$, to succinctly represent linear combinations with non-negative integer coefficients. For \tabedmtcp{}, we sample $W_t = [w_t]$, where $w_t = \text{softmax}([1/{e_t^{(1)}}, \ldots, 1/{e_t^{(m)}}]; \tau) \in \mathbb{R}^{m}$ using soft labels, temperature $\tau$, and $n=1$, offering an efficient alternative as $m$ increases. We set temperature $\tau=1$ by default.

\subsection{Draft block length $\gamma$}
\label{sec:app_gamma}

To evaluate the effect of draft block length $\gamma$, we conducted additional experiments with draft lengths of $3$ and $7$, extending our baseline setting of $\gamma=5$. 
As shown in~\cref{tab:gamma-3,tab:gamma-7}, the overall trends remain consistent across different values of $\gamma$, demonstrating the robustness of our method to the choice of draft block length. 
Specifically, the dynamic ensemble drafting methods, \oursnotation{} and \oursmtcpnotation{}, consistently outperform or closely match the best-performing baselines across the most of tested configurations, confirming that our findings are robust to this hyperparameter.



\begin{table*}[t]
\centering
\sisetup{table-auto-round}
\renewcommand{\arraystretch}{1.0}
\captionsetup{skip=5pt}
\caption{
Block efficiency across datasets for $\gamma=3$. \best{Bold} and \secondbest{underlined} denote the best and second-best results.
}
\label{tab:gamma-3}
\resizebox{1.0\textwidth}{!}{
\begin{tabular}{ll|ccccccc||cc}
\toprule
$\gamma=3$ & Drafting &
\multicolumn{7}{c||}{\titlecap{\standard{} datasets}} &
\multicolumn{2}{c}{\titlecap{\ood{} datasets}} \\
\cmidrule(lr){1-2}\cmidrule(lr){3-9}\cmidrule(lr){10-11}
Type & Method & LLaVA\,-W & DocVQA & POPE & MMVet & IEdit & MB & Spot & PSV & VIST \\
\midrule
\multirow{4}{*}{\singleabbv{}} 
& M & 2.09 & \secondbest{2.03} & \best{2.24} & \best{2.06} & 2.06 & 1.85 & 2.15 & 1.19 & 1.16 \\
& T & 2.04 & 1.97 & 2.11 & 2.02 & 2.06 & 2.14 & 2.11 & 1.95 & 1.94 \\
& C & 2.08 & 2.01 & \secondbest{2.23} & 2.02 & \secondbest{2.11} & 2.17 & 2.12 & \secondbest{1.98} & \secondbest{1.99} \\
& P & 2.07 & 1.97 & 2.16 & 2.03 & 2.07 & 2.10 & 2.07 & 1.97 & \secondbest{1.99} \\
\cmidrule(lr){1-11} 
\multirow{2}{*}{\ensembleabbv{}} 
& \textbf{\oursnotation{}}   & \best{2.10} & \best{2.04} & 2.22 & {2.05} & \secondbest{2.11} & \secondbest{2.19} & \best{2.17} & 1.93 & 1.93 \\
& \textbf{\oursmtcpnotation{}} & \best{2.10} & 2.02 & 2.22 & \best{2.06} & \best{2.14} & \best{2.20} & \best{2.17} & \best{2.01} & \best{2.03} \\
\bottomrule
\end{tabular}
}
\end{table*}

\begin{table*}[t]
\centering
\sisetup{table-auto-round}
\renewcommand{\arraystretch}{1.0}
\captionsetup{skip=5pt}
\caption{
Block efficiency across datasets for $\gamma=7$. \best{Bold} and \secondbest{underlined} denote the best and second-best results.
}
\label{tab:gamma-7}
\resizebox{1.0\textwidth}{!}{
\begin{tabular}{ll|ccccccc||cc}
\toprule
$\gamma=7$ & Drafting &
\multicolumn{7}{c||}{\titlecap{\standard{} datasets}} &
\multicolumn{2}{c}{\titlecap{\ood{} datasets}} \\
\cmidrule(lr){1-2}\cmidrule(lr){3-9}\cmidrule(lr){10-11}
Type & Method & LLaVA\,-W & DocVQA & POPE & MMVet & IEdit & MB & Spot & PSV & VIST \\
\midrule
\multirow{4}{*}{\singleabbv{}} 
& M & 2.31 & \secondbest{2.22} & \best{2.59} & \secondbest{2.24} & 2.26 & 2.01 & 2.41 & 1.19 & 1.16 \\
& T & 2.21 & 2.14 & 2.34 & 2.19 & 2.28 & 2.42 & 2.34 & 2.05 & 2.05 \\
& C & 2.26 & 2.19 & \secondbest{2.56} & 2.21 & \secondbest{2.35} & \secondbest{2.48} & 2.37 & \secondbest{2.11} & \secondbest{2.12} \\
& P & 2.26 & 2.12 & 2.45 & 2.21 & 2.27 & 2.35 & 2.28 & {2.08} & \secondbest{2.12} \\
\cmidrule(lr){1-11} 
\multirow{2}{*}{\ensembleabbv{}} 
& \textbf{\oursnotation{}}   & \best{2.32} & \best{2.23} & {2.54} & \secondbest{2.24} & 2.34 & \secondbest{2.48} & \secondbest{2.44} & 2.03 & 2.04 \\
& \textbf{\oursmtcpnotation{}} & \best{2.32} & 2.21 & 2.55 & \best{2.25} & \best{2.39} & \best{2.51} & \best{2.47} & \best{2.15} & \best{2.19} \\
\bottomrule
\end{tabular}
}
\end{table*}

\subsection{TABED-AdaBoost: A Theoretically Grounded Variant}
\label{sec:app_adaboost}

To complement our empirical ablations with a more theoretically grounded alternative, we study a multi-class AdaBoost-inspired weighting scheme ({TABED-AdaBoost}), which assigns a fixed ensemble weight to each drafting method based on its error rate.
Concretely, each drafting method $m$ is assigned
\begin{equation}
w_m = \ln\!\left(\frac{1}{\epsilon_m}-1\right) + C\cdot\ln(K-1),
\end{equation}
where $\epsilon_m$ denotes the method's error rate, $K$ is the vocabulary size, and $C$ is a constant.

Table~\ref{tab:adaboost} compares \oursnotation{}-AdaBoost with the original \oursnotation{}.
Across five of the seven in-domain benchmarks, \oursnotation{} performs comparably to \oursnotation{}-AdaBoost, while achieving better robustness on OOD datasets (PororoSV and VIST), suggesting that \oursnotation{}'s adaptive weighting quickly upweights the stronger drafting method.

\begin{table*}[t]
  \centering
  \sisetup{table-auto-round,mode=text}
  \renewcommand{\arraystretch}{0.85}
  \small

  \caption{Comparison between TABED-AdaBoost and TABED.}
  \vspace{-0.5em}
  \label{tab:adaboost}

  {%
    \setlength{\tabcolsep}{3.2pt}
    \begin{tabular}{lccccccccc}
      \toprule
      & {LLaVA-W} & {DocVQA} & {POPE} & {MMVet} & {IEdit} & {MB} & {Spot} & {PSV} & {VIST} \\
      \midrule
      {\textbf{\oursnotation{}-AdaBoost}} & 2.28 & 2.12 & 2.48 & 2.23 & 2.26 & 2.28 & 2.33 & 1.96 & 1.87 \\
      {\textbf{\oursnotation{}}}          & 2.26 & 2.16 & 2.52 & 2.21 & 2.29 & 2.39 & 2.36 & 2.02 & 2.04 \\
      \bottomrule
    \end{tabular}
  }%

  \vspace{-1em}
\end{table*}

\section{More Qualitative Samples of Dynamic Ensemble Weights}
\label{app:quali_w}
\begin{figure}[t]
    \centering
    \includegraphics[width=\linewidth]{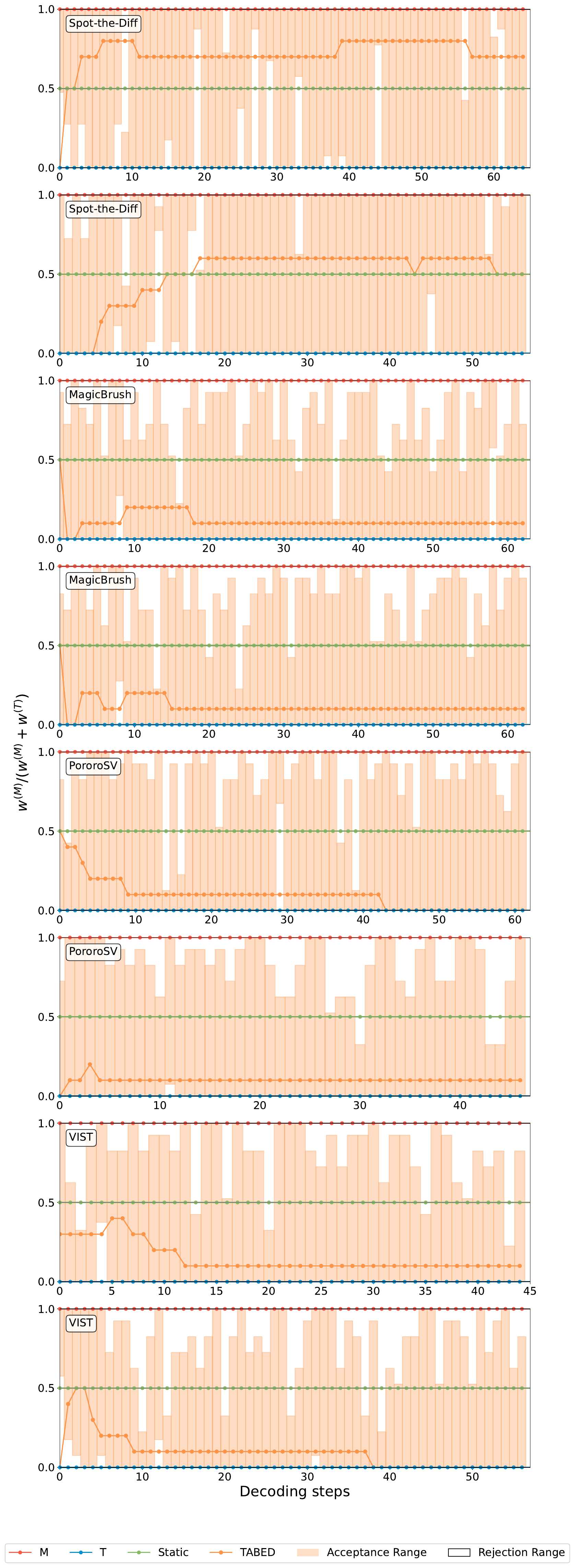}
    \caption{Additional qualitative samples of dynamic ensemble weights}
    \label{fig:quali_adaptive_weights_apd}
\end{figure}

In this section, we provide further qualitative examples of dynamic ensemble weights. In~\cref{fig:quali_adaptive_weights_apd}, the y-axis denotes the proportion of $w^{(M)}$ relative to $w^{(M)}+w^{(T)}$, and the x-axis indicates decoding steps. In comparison to other single drafting methods, M and T ($y=1$ and $0$), or the static ensembling method MT ($y=0.5$), adaptive ensemble weights $w^*$ predicted by \oursadaens{} (the blue graph) effectively explore the region shaded, representing the weights that lead to token acceptance by the target model.

\section{Full Results for \ours{} with Token Tree Verification}
\label{app:token-tree}
In this section, we present the full results corresponding to the main experiments in~\cref{tab:width-avg} for all tree widths $\treewidth$ evaluated in this paper, including both single (multimodal (M) and text-only (T)) and dynamic ensemble approaches (\oursnotation{}). As shown in~\cref{tab:full-width-2,tab:full-width-3}, the findings from~\cref{sec:study,sec:ensemble} remain consistent when combined with orthogonal enhanced verification methods across different tree widths $\treewidth$.

\begin{table*}[!ht]
\centering
\sisetup{table-auto-round}
\caption{
Block efficiency across datasets for \textbf{$\treewidth=2$}. 
\best{Bold} and \secondbest{underlined} denote the best and second-best results.
}
\label{tab:full-width-2}
\resizebox{1.0\textwidth}{!}
{
\begin{tabular}{ll|cccccccc||cc}
\toprule
$\treewidth=2$ & Drafting &
\multicolumn{8}{c||}{\titlecap{\standard{} datasets}} &
\multicolumn{2}{c}{\titlecap{\ood{} datasets}} \\
\cmidrule(lr){1-2}\cmidrule(lr){3-10}\cmidrule(lr){11-12}
Type & Method & LLaVA-W & DocVQA & POPE & MMVet & IEdit & MB & Spot & Avg. & PSV & VIST \\
\midrule
\multirow{2}{*}{\singleabbv{}} 
& M & \best{2.93} & \secondbest{2.72} & \best{3.48} & \best{2.82} & 2.77 & 2.37 & \secondbest{3.13} & \secondbest{2.89} & 1.32 & 1.27 \\
& T & 2.82 & 2.64 & 2.92 & 2.76 & \secondbest{2.84} & \secondbest{2.90} & 3.04 & 2.85 & \best{2.70} & \best{2.73} \\
\cmidrule(lr){1-12} 
\multirow{1}{*}{\ensembleabbv{}} 
& \textbf{\oursnotation{}} & \secondbest{2.92} & \best{2.75} & \secondbest{3.39} & \best{2.82} & \best{2.91} & \best{2.97} & \best{3.17} & \best{2.99} & \secondbest{2.62} & \secondbest{2.66} \\
\bottomrule
\end{tabular}
}
\end{table*}

\begin{table*}[!ht]
\centering
\sisetup{table-auto-round}
\caption{
Block efficiency across datasets for \textbf{$\treewidth=3$}. 
\best{Bold} and \secondbest{underlined} denote the best and second-best results.
}
\label{tab:full-width-3}
\resizebox{1.0\textwidth}{!}{
\begin{tabular}{ll|cccccccc||cc}
\toprule
$\treewidth=3$ & Drafting &
\multicolumn{8}{c||}{\titlecap{\standard{} datasets}} &
\multicolumn{2}{c}{\titlecap{\ood{} datasets}} \\
\cmidrule(lr){1-2}\cmidrule(lr){3-10}\cmidrule(lr){11-12}
Type & Method & LLaVA-W & DocVQA & POPE & MMVet & IEdit & MB & Spot & Avg. & PSV & VIST \\
\midrule
\multirow{2}{*}{\singleabbv{}} 
& M & \secondbest{3.31} & \secondbest{3.10} & \best{3.98} & \best{3.22} & 3.08 & 2.66 & \secondbest{3.63} & \secondbest{3.28} & 1.41 & 1.35 \\
& T & 3.22 & 2.98 & 3.34 & 3.10 & \secondbest{3.20} & \secondbest{3.27} & 3.54 & 3.24 & \best{3.16} & \best{3.21} \\
\cmidrule(lr){1-12} 
\multirow{1}{*}{\ensembleabbv{}} 
& \textbf{\oursnotation{}} & \best{3.32} & \best{3.11} & \secondbest{3.87} & \best{3.22} & \best{3.25} & \best{3.33} & \best{3.64} & \best{3.39} & \secondbest{3.03} & \secondbest{3.14} \\
\bottomrule
\end{tabular}
}
\end{table*}

\section{Details for Caption Drafting}
\label{app:caption}
\begin{table}[h]
\centering
\sisetup{table-auto-round}
\resizebox{1\linewidth}{!}{%
\begin{tabular}{c c | c c c}
\toprule
& & \multicolumn{3}{c}{Latency (s)} \\
\cmidrule(lr){3-5}
Model & Type & $n=1$ &$n=2$&$n=5$\\
\midrule
\multirow{1}{*}{\shortstack[l]{Target LVLM (\prefill{})}}&LLaVA-1.5 7B & 0.112 &0.207& 0.540\\
\midrule
\multirow{2}{*}{\shortstack[l]{Image Captioning}}&BLIP & \best{0.054} & \best{0.055}& \best{0.074}\\
&Florence-2 & \secondbest{0.105} & \secondbest{0.149} & \secondbest{0.292} \\
\bottomrule
\end{tabular}
}
\caption{Latency analysis of image captioning models. BLIP and Florence-2 captioning latencies are lower than the target LVLM's prefilling latency. Parallel processing can therefore hide captioning latency without affecting time to first token.}
\label{tab:caption_latency}
\end{table}
In this section, we describe various types of lightweight image captioning models that can be used for caption drafting (\cref{app:cap_list}). We then demonstrate that captioning model inference completes earlier than the target model's prefilling by analyzing the captioning model's latency (\cref{sec:app_cap_latency}).


\subsection{Captioning Models}
\label{app:cap_list}

\paragraph{BLIP \citep{li2022blip}}
A vision-language model trained on bootstrapped synthetic captions. It uses a visual transformer and the text encoder of BERT \cite{devlin-etal-2019-bert} to separately encode image and text.

\url{https://huggingface.co/Salesforce/blip-image-captioning-base}
    
\paragraph{BLIP-2 \citep{li2023blip}}
A vision-language model using a frozen off-the-shelf image encoder and LLM. A querying transformer trained using boostrapped data is included for cross-modal alignment.

\url{https://huggingface.co/Salesforce/blip2-opt-2.7b}

\paragraph{Florence-2 \citep{xiao2024florence}}
A vision-language model that is instruction-trained for a variety of tasks. Its architecture consists of a single sequence-to-sequence transformer and a vision encoder.

\url{https://huggingface.co/microsoft/Florence-2-large-ft}

\subsection{Latency Analysis}
\label{sec:app_cap_latency}
It is important to ensure that the captioning model runs fast enough so that it does not delay drafting. In this line, we measure in \cref{tab:caption_latency} the time taken by the two captioning models, BLIP and Florence-2, to generate captions. The results demonstrate captioning completes earlier than target model's prefilling.
\section{Evaluation of Target Model}
\label{app:samples}

\begin{table}[h]
\centering
\sisetup{table-auto-round}
\resizebox{1.0\linewidth}{!}{%
\begin{tabular}{c|rrr}
\toprule
\multicolumn{1}{c}{\multirow{2}{*}{Model}} & \multicolumn{2}{c}{\titlecap{datasets}} \\
\cmidrule(lr){2-3}
 &  Spot-the-Diff & MagicBrush  \\
\midrule
 GPT-4o  & 19.1 & 10.5 \\
 GPT-4o-mini & 12.0 & 9.1 \\
 LLaVA-1.5 7B & 13.9 & 9.6 \\
 LLaVA-1.5 7B (text-only) & 11.1 & 8.4 \\
\bottomrule
\end{tabular}
}
\caption{
Evaluation results for the target model on benchmark datasets. 
Our target model, LLaVA-1.5 7B, demonstrates performance comparable to GPT-4o-mini. 
The injection of multimodal information in the input scenarios of the benchmark datasets benefits the target model, LLaVA-1.5 7B.
}
\label{tab:app_eval2}
\end{table}

In this section, we present both quantitative results (\cref{sec:target_quant_eval}) and qualitative samples (\cref{sec:target_quali_eval}) produced by LLaVA-1.5 7B, which serves as the target model in our experimental setting.

\subsection{Quantitaive Evaluation}
\label{sec:target_quant_eval}
We assess the output quality of LLaVA-1.5 7B using benchmark datasets, as shown in \cref{tab:app_eval2}. The results demonstrate that the quality of LLaVA-1.5 7B is comparable to GPT-4o-mini.
The injection of multimodal information benefits the target model, LLaVA-1.5 7B, from the perspective of output quality.

\subsection{Generated Samples}
\label{sec:target_quali_eval}
\begin{figure*}[!htb]
\centering
    \begin{subfigure}[b]{0.32\linewidth}
        \centering
        \resizebox{\linewidth}{!}{%
\includegraphics[height=3.6cm]{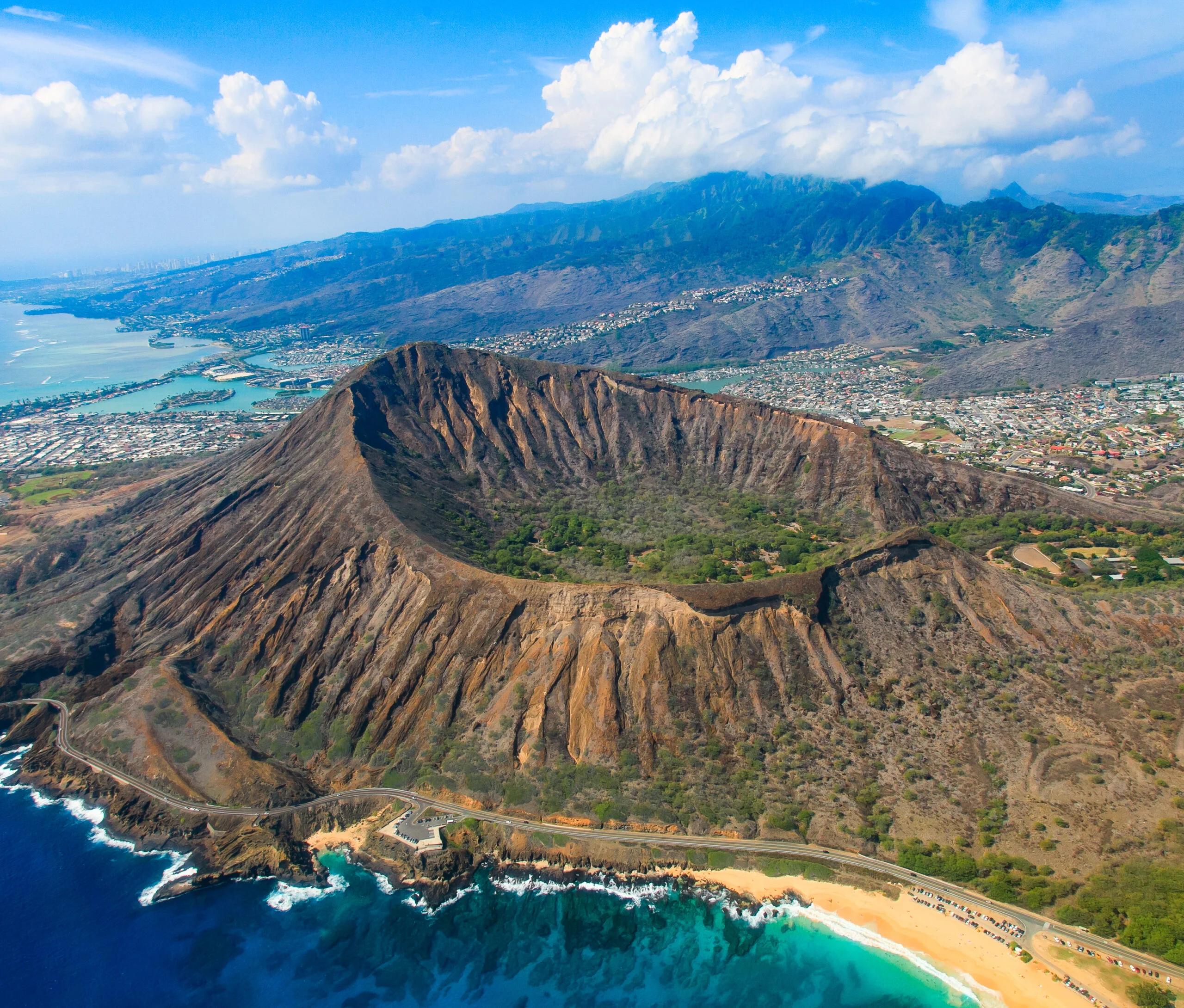}
}
        \caption{LLaVA-Bench (In-the-Wild)}
        \label{ex:chartqa}
    \end{subfigure}
    \hfill
    \begin{subfigure}[b]{0.32\linewidth}
        \centering
        \resizebox{\linewidth}{!}{%
\includegraphics[height=3.6cm]{acl/latex/figures/assets/docvqa_sample.jpg}
        }
        \caption{DocVQA}
        \label{ex:textvqa}
    \end{subfigure}
    \hfill
    \begin{subfigure}[b]{0.32\linewidth}
        \centering
        \resizebox{\linewidth}{!}{%
\includegraphics[height=3.6cm]{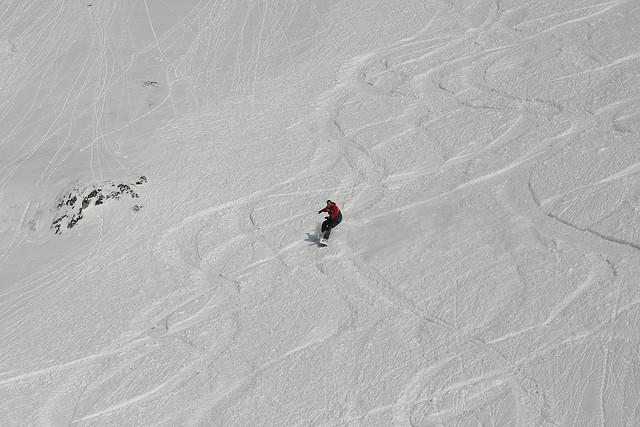}
        }
        \caption{POPE}
        \label{ex:vqav2}
    \end{subfigure}
        \hfill

\bigskip
    \begin{subfigure}[b]{0.32\linewidth}
        \centering
        \resizebox{\linewidth}{!}{%
\includegraphics[height=3.6cm]{acl/latex/figures/assets/mmvet_sample.png}
        }
        \caption{MMVET}
        \label{ex:hallusion}
    \end{subfigure}
    \begin{subfigure}[b]{0.32\linewidth}
        \centering
        \resizebox{\linewidth}{!}{%
\includegraphics[height=3.6cm]{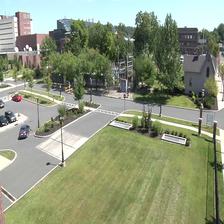}
\includegraphics[height=3.6cm]{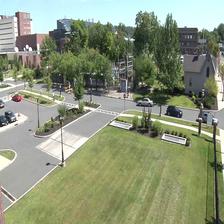}
        }
        \caption{Spot the Difference}
        \label{ex:spot}
    \end{subfigure}
    \begin{subfigure}[b]{0.32\linewidth}
        \centering
        \resizebox{\linewidth}{!}{%
\includegraphics[height=3.6cm]{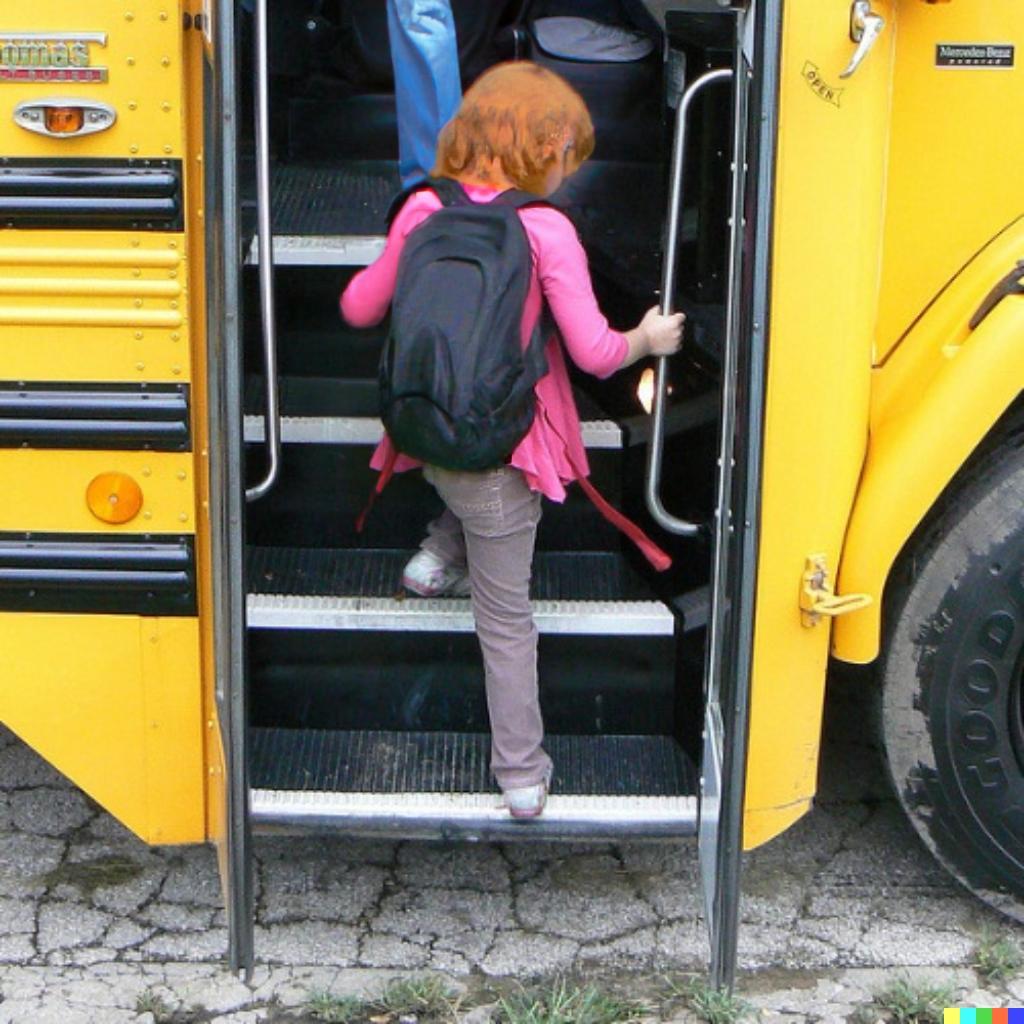}
\includegraphics[height=3.6cm]{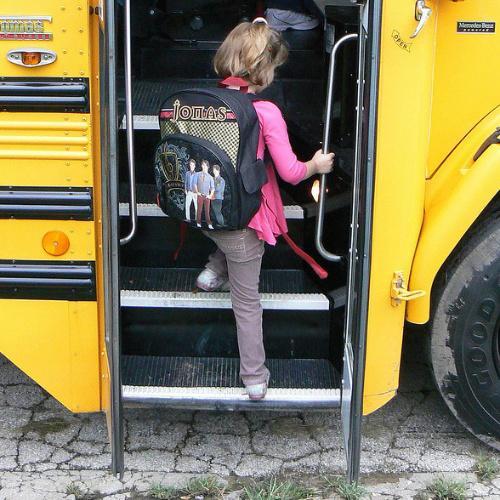}
        }
        \caption{MagicBrush}
        \label{ex:mb}
    \end{subfigure}

\bigskip

    \begin{subfigure}[b]{1\linewidth}
        \centering
        \resizebox{\linewidth}{!}{%
\includegraphics[height=3.6cm]{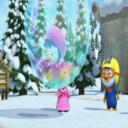}
\includegraphics[height=3.6cm]{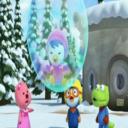}
\includegraphics[height=3.6cm]{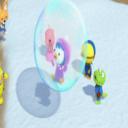}
\includegraphics[height=3.6cm]{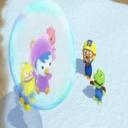}
\includegraphics[height=3.6cm]{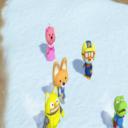}
        }
        \caption{PororoSV}
        \label{ex:psv}
    \end{subfigure}

\bigskip

    \begin{subfigure}[b]{1\linewidth}
        \centering
        \resizebox{\linewidth}{!}{%
\includegraphics[height=3.6cm]{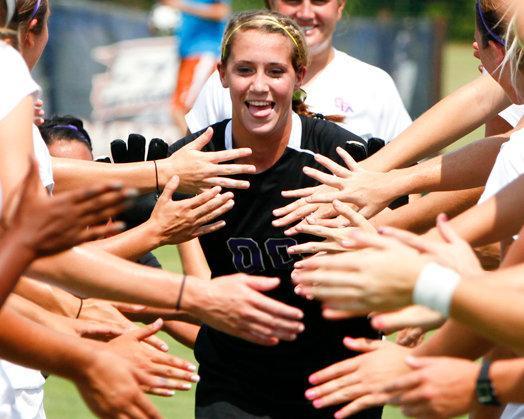}
\includegraphics[height=3.6cm]{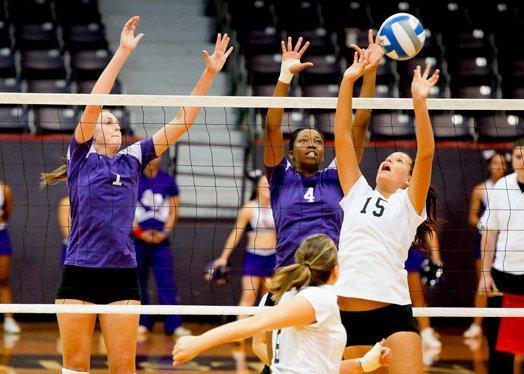}
\includegraphics[height=3.6cm]{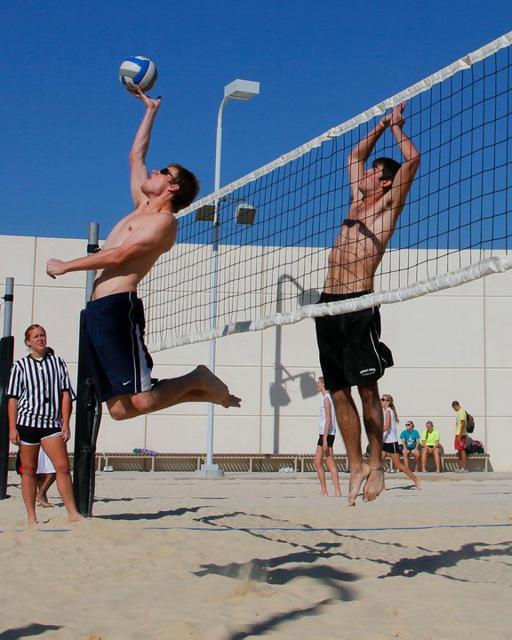}
\includegraphics[height=3.6cm]{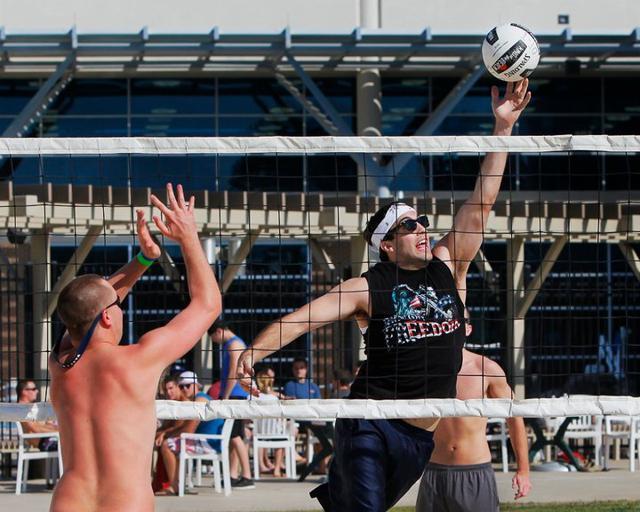}
\includegraphics[height=3.6cm]{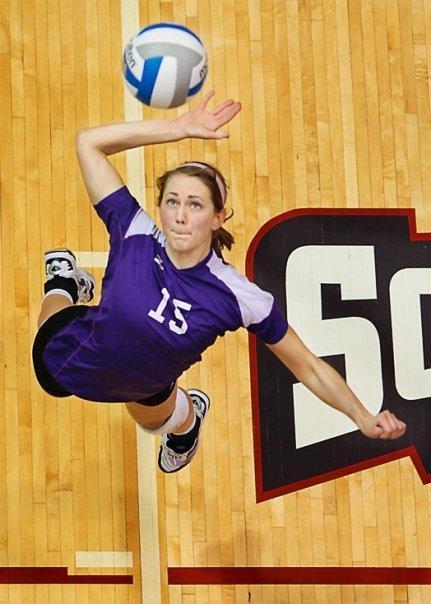}
        }
        \caption{VIST}
        \label{ex:vist}
    \end{subfigure}
    
    \caption{Qualitative image samples of benchmark and OOD datasets. The corresponding questions and answers are presented in \cref{app:samples}.}
    \label{fig:app_dataset_1}
\end{figure*}

\paragraph{Spot The Difference} [\cref{ex:spot}]

\textit{Q:} What has changed from the first image to the second?

\textit{A:} The image shows a city street with a grassy area in the middle of the road. In the first image, there are several cars parked along the street, and a few people are walking around. The street appears to be relatively empty, with no traffic or significant activity. In the second image, the street is bustling with activity. There are numerous cars parked along the street, and several people are walking around. The scene is more lively and dynamic, with a higher level of activity compared to the first image.

\paragraph{MagicBrush} [\cref{ex:mb}]

\textit{Q:} Please provide the image edit instruction that can transfrom the source image to the target image.

\textit{A:} To transform the source image into the target image, follow these steps: 1. Remove the red-haired girl from the scene, as she is not part of the target image. 2. Replace the girl with a back pack, which should be placed on the stairs. 3. Add a bus in the background, making sure it is yellow and occupies the entire width of the image. 4. Place a person in the background, standing near the bus, to give a sense of scale and context. 5. Adjust the lighting and shadows to create a more realistic scene.

\paragraph{PororoSV} [\cref{ex:psv}]

\textit{Q:} Taking the stories from the initial images into account, can you conclude the story using the elements of the final image?

\textit{A:} The image features a group of cartoon characters, including a penguin, a fox, and a pink character, standing in a snowy environment. The characters are positioned in various locations, with some standing closer to the foreground and others further back. The scene appears to be a part of a video game or animated series, as the characters are interacting with each other and their surroundings.

\paragraph{VIST} [\cref{ex:vist}]

\textit{Q:} With the narratives paired with the initial images, how would you conclude the story using the last picture?

\textit{A:} The image captures a thrilling moment during a volley ball game, with a woman leaping into the air to hit the ball. She is wearing a purple shirt and is in the process of jumping to make contact with the sports ball. The scene is filled with excitement and anticipation as the players on the court watch her performance. The volley ball game is taking place on a court, and the players are actively engaged in the match.

\section{Empirical Validation of Latency and Throughput Remarks}
\label{app:remarks}

In this section, we present detailed empirical analyses that support the theoretical assumptions introduced in the main text. 
We first report measurements of the draft-to-target latency ratio $T_q/T_p$ across representative LVLM benchmarks as well as a detailed derivation of the walltime speedup (\cref{sec:app_latency_ratio}).

We then validate the three latency invariance remarks formulated in \cref{remark:gamma,remark:length,remark:batch} and discuss their implications for speculative decoding efficiency (\cref{sec:app_remarks_validation}). 
Finally, we analyze throughput and computational cost to confirm that the additional compute introduced by \ours{} is negligible in practice (\cref{sec:app_throughput}).

\subsection{Empirical Measurement of Draft-to-Target Latency Ratio $T_q/T_p$}
\label{sec:app_latency_ratio}
We empirically measure the draft-to-target latency ratio ($T_q/T_p$) across well-known LVLM benchmarks~\citep{li2024llavanextinterleavetacklingmultiimagevideo}. 
This ratio allows us to estimate the expected walltime speedup by substituting it into \cref{eq:speedup}. 
All measurements reported in \cref{tab:latency_ratio} are averaged across datasets. 

This allows us to calculate the expected speedup by inputting the draft-to-target latency ratio into \cref{eq:speedup}. 
Specifically, the draft-to-target latency ratio between the 7B target model and the 68M draft model is measured as $0.063$~\cref{tab:latency_ratio}. The average block efficiency of \ours{} is $2.29$, computed as the mean of $2.31$ and $2.27$, and the block length $\gamma$ used in~\cref{tab:acl_main} is set to $5$. Substituting these values into~\cref{eq:speedup} yields an estimated wall-time speedup of $\text{speedup} = {2.29}/({5 \times 0.063 + 1}) = 1.74$.
We report speedup in this factorized form (block efficiency, draft-to-target latency ratio, and block length) because it is more robust to measurement noise.

\begin{table}[h]
\centering
\sisetup{table-auto-round}
\resizebox{\linewidth}{!}{
\begin{tabular}{c c | c}
\toprule
$\targetmodel$ Size & $\draftmodel$ Size & draft-to-target latency ratio $T_q/T_p$ \\
\midrule
\multirow{2}{*}{7B} & 68M & 0.063 \\
                    & 160M & 0.206 \\
\midrule
\multirow{2}{*}{13B} & 68M & \best{0.042} \\
                     & 160M & 0.137 \\
\bottomrule
\end{tabular}
}
\caption{Empirical measurements of the draft-to-target latency ratio ($T_q/T_p$), covering all model sizes.
All measurements are averaged across datasets.
}
\label{tab:latency_ratio}
\end{table}

\subsection{Analysis of Latency in \ours{}}
\label{sec:app_remarks_validation}
\begin{figure*}[t]
    \centering
    \includegraphics[width=\linewidth]{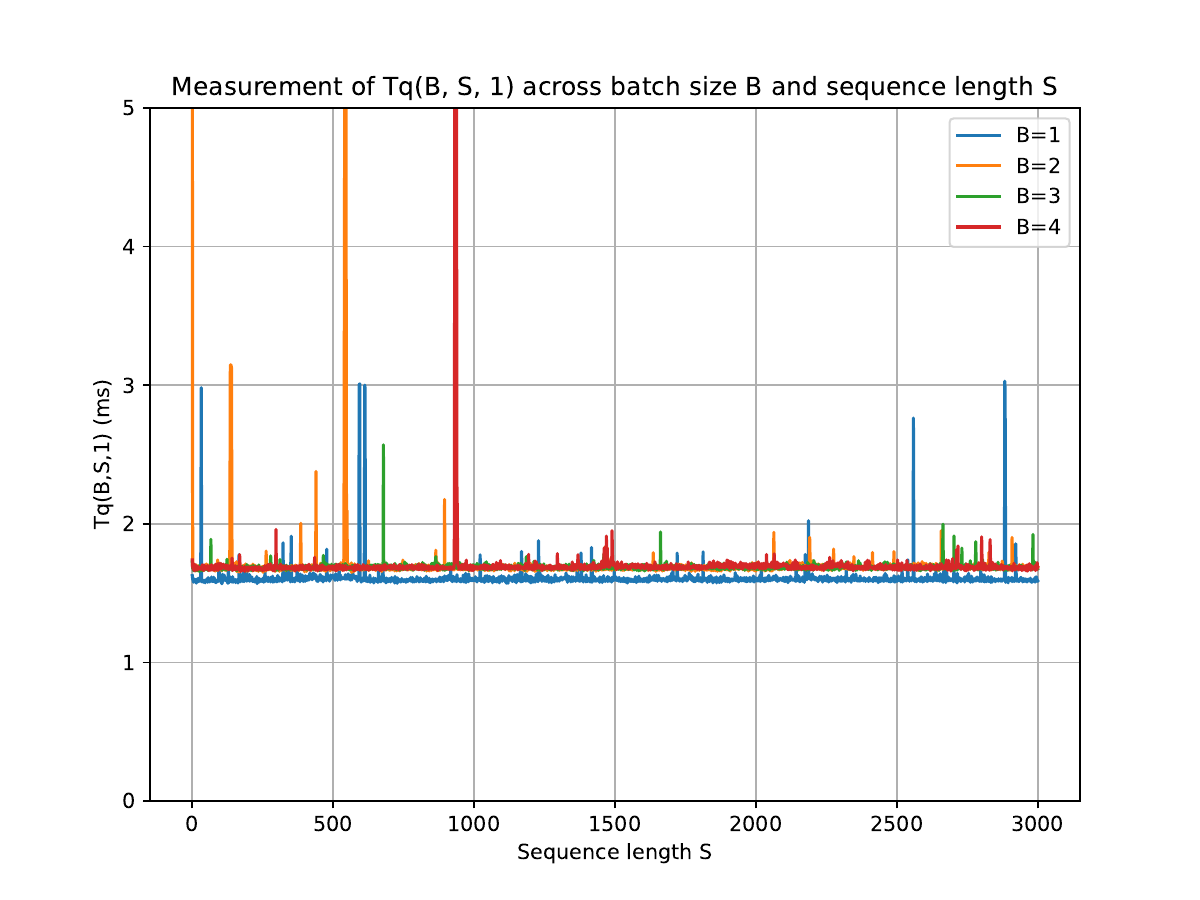}
\caption{Empirical demonstration of Remarks 2 and 3.}
\label{fig:remark_plot_23}
\end{figure*}
\begin{figure*}[t]
    \centering
    \includegraphics[width=\linewidth]{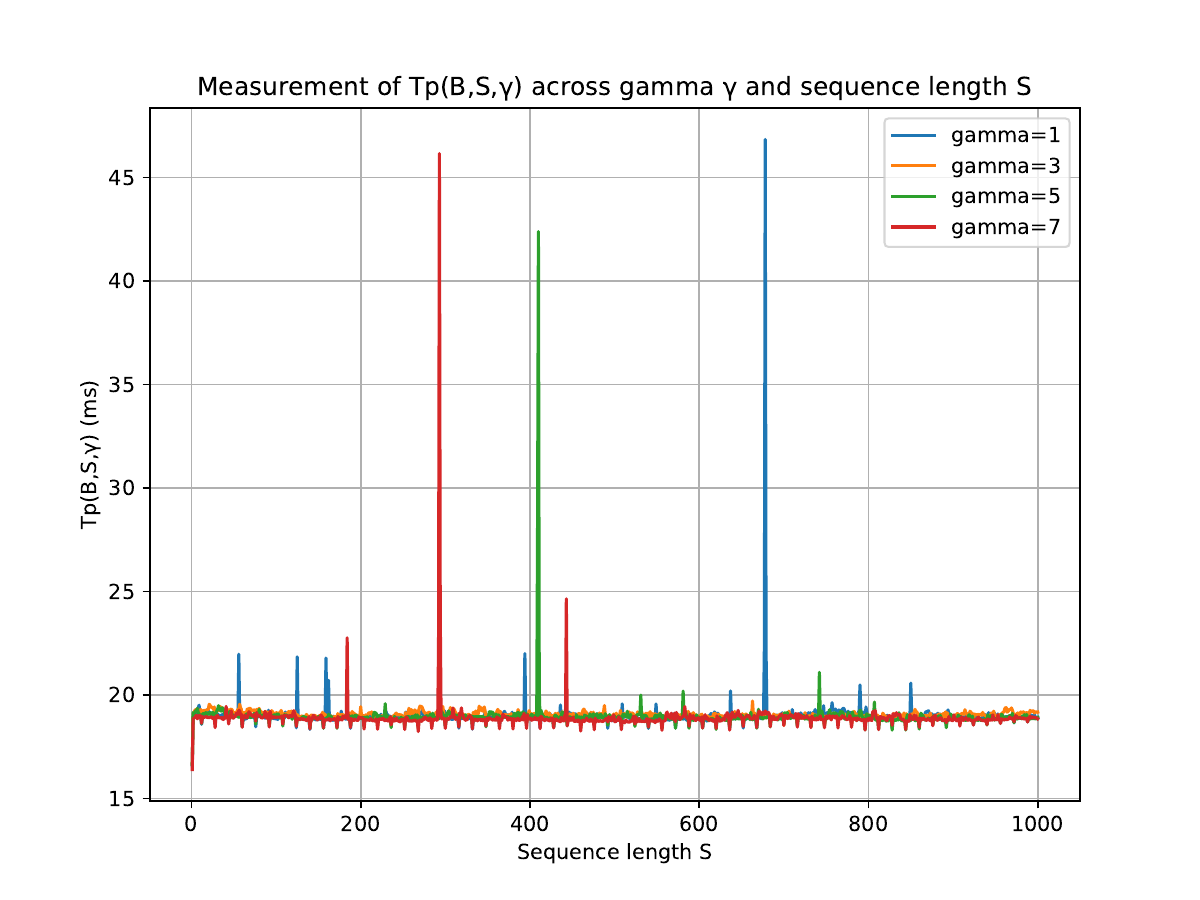}
\caption{Empirical demonstration of Remark 1.}
\label{fig:remark_plot_1}
\end{figure*}

To further analyze and discuss the effects of the \ours{} when latency ratio ${T_q}/{T_p}$ is small like shown in ~\cref{tab:latency_ratio} to the latency and throughput of speculative decoding system, we formulate the remarks~\cref{remark:gamma,remark:length,remark:batch} and empirically showed that the effect is negligible. 

Focusing on the case where the remarks apply to speculative decoding settings, we use LLaVA-1.5 7B and LLaVA-1.5 68M to measure $T_p(B,S,\gamma)$ for \cref{remark:gamma} and $T_q(B,S,1)$ for \cref{remark:batch,remark:length}, respectively.
Also, we set  $S$ (\eg, $S \leq 3$k) and sufficiently small $B$ (\eg, $B \leq 4$) and $\gamma$ (\eg, $\gamma \leq 10$) to cover all configurations used throughout the paper.
All experiments are conducted on an A100 80GB GPU using the fp16 data type for the models.
As a result, the decoding phase displays the following observations:

\begin{remark}
\label{remark:gamma}
For given $B$ and $S$, regardless of $\gamma$, $T(B,S,\gamma)$ remains approximately constant (\eg, $|\Delta T/T| < 0.05$).
\end{remark}
\begin{remark}
\label{remark:length}
For a given $B$, regardless of $S$, $T(B,S,1)$ remains approximately constant (\eg, $|\Delta T/T| < 0.05$).
\end{remark}
\begin{remark}
\label{remark:batch}
For a given $S$, regardless of $B$, $T(B,S,1)$ remains approximately constant (\eg, $|\Delta T/T| < 0.05$).
\end{remark}

\cref{fig:remark_plot_23} shows $T_q(B, S, 1)$ in milliseconds for sequence lengths up to 3k for each batch size $B \in \lbrace 1,2,3,4 \rbrace$. For moderate sequence lengths $S \leq 3\text{k}$, $T_q$ varies by no more than 5\% for each $B$, which supports Remark 2. Similarly, when comparing different $B$s with a fixed $S$, $T_q$ varies by no more than 5\%, which supports Remark 3.


\cref{fig:remark_plot_1} shows $T_p(B, S, \gamma)$ in milliseconds for each $\gamma\in\lbrace 1,3,5,7 \rbrace$. We test the case of B=1, which aligns with our experimental settings where the target model always performs inference on a single batch. Over the values of $\gamma$ considered, $T_p$ varies by no more than 5\%.

\subsection{Analysis of Throughput in \ours{}}
\label{sec:app_throughput}
We further evaluate the effect of \ours{} from a throughput perspective rather than latency. 
Because $\draftmodel$ is substantially smaller than $\targetmodel$, batch inference using $\draftmodel$ (68M) introduces negligible computational overhead relative to the total workload dominated by $\targetmodel$ (7B). 
Prior studies have shown that Transformer decoding is memory-bound rather than compute-bound~\citep{chen2024magicdec,fu2024challenges}, implying that such additional compute does not meaningfully affect walltime. 

\cref{tab:gflops} presents the FLOPs analysis supporting this claim: the additional 0.08 GFLOPs per prediction (0.17–0.09) from $\draftmodel$ is minimal compared to the 66.08 GFLOPs required by $\targetmodel$. 
All measurements are obtained using a single image (576 projected image tokens) and 24 text tokens. 
This confirms that \ours{} introduces no meaningful throughput degradation in practice.
\begin{table}[!h]
\centering
\caption{Computational cost (GFLOPs) of the target model ($\targetmodel{}$) and the draft model ($\draftmodel{}$) under different drafting methods (Multimodal and \oursnotation{}).}
\label{tab:gflops}
\resizebox{\linewidth}{!}{
\begin{tabular}{lccc}
\toprule
\textbf{Model} & {$\targetmodel$} & {$\draftmodel{}$ (Multimodal)} & $\draftmodel{}$ (\textbf{\oursnotation{}}) \\
\midrule
\textbf{GFLOPs} & 66.08 & 0.09 & 0.17 \\
\bottomrule
\end{tabular}
}
\end{table}


\section{Training and Evaluation of Draft Models}
\label{app:training}
In this section, we present a more detailed overview of our custom training procedure for the draft models (\cref{sec:app_train_detail}).
We then evaluate our primary draft model, LLaVA-1.5 68M, on multimodal tasks to ensure it has the capability to properly perceive multimodality, and we provide some qualitative samples from the draft model (\cref{sec:app_draft_evaluation}).

\subsection{Details of Training}
\label{sec:app_train_detail}

\paragraph{LLaVA-1.5~\citep{liu2024improved}}
 
The process for developing draft models with the LLaVA-1.5 (68M, 160M) training recipe was divided into two stages: pre-training and instruction fine-tuning (IFT). 
Pre-training focused on training the projector while the parameters of the LLM and vision encoder were frozen. During the IFT stage, visual instruction tuning was used to teach the LLM to follow multimodal instructions. The vision encoder remained frozen throughout both stages.
The hyperparameters used for each stage are described in \cref{tab:hyperparameters_1.5}.
We trained the draft model using datasets curated by the original author of LLaVA-1.5. For more training details, see \href{link}{https://github.com/haotian-liu/LLaVA/tree/main}.
\begin{table}[ht]
\centering
\begin{subtable}[t]{0.48\linewidth}
\centering
\resizebox{\linewidth}{!}{
\sisetup{table-auto-round}
\begin{tabular}{ c c }
\toprule
\textbf{Hyperparameter} & \textbf{Value} \\
\midrule
Training Epochs & 1 \\
Batch Size & 256 \\
Learning Rate (LR) & 1e-3 \\
LR Schedule Type & Cosine\\
Warm-up Ratio & 0.03 \\
Weight Decay & 0.0 \\
\bottomrule
\end{tabular}
}
\caption{Pretraining stage}
\label{tab:app_pretraining_hyperparameters}
\end{subtable}
\hfill
\begin{subtable}[t]{0.48\linewidth}
\centering
\resizebox{\linewidth}{!}{
\sisetup{table-auto-round}
\begin{tabular}{ c c }
\toprule
\textbf{Hyperparameter} & \textbf{Value} \\
\midrule
Training Epochs & 1 \\
Batch Size & 128 \\
Learning Rate (LR) & 2e-5 \\
LR Schedule Type & Cosine\\
Warm-up Ratio & 0.03 \\
Weight Decay & 0.0 \\
\bottomrule
\end{tabular}
}
\caption{Instruction fine-tuning stage}
\label{tab:app__finetuning_hyperparameters}
\end{subtable}
\caption{Details of hyperparameters used in LLaVA-1.5 training}
\label{tab:hyperparameters_1.5}
\end{table}

\paragraph{LLaVA-OneVision~\citep{li2024llava-onevision}}
The development of draft models using the LLaVA-OneVision (LLaVA-OV) training recipe was divided into three stages: language-image alignment, high-quality knowledge learning, and visual instruction tuning. 
In the language-image alignment stage, visual features were aligned with the word embedding space of LLMs. 
High-quality knowledge learning balanced computational efficiency with the integration of new knowledge into LVLMs. 
Visual instruction tuning consisted of two phases: (i) Single-Image Training, where the model learned to perform visual tasks using instructions from single images, and (ii) OneVision Training, where the model learned to execute multi-image visual tasks using a blend of video, single-image, and multi-image data.
During the language-image alignment stage, only the projector for aligning visual features was updated, whereas all components including LLM were updated in the following three stages.
We trained the draft model using datasets curated by the original author of LLaVA-OV~\citep{li2024llava-onevision}. 
The hyperparameters used for each stage are described in \cref{tab:hyperparameters_ov}, and the learning rate for the vision encoder is one-fifth of that for the LLM across all stages.
For more details, visit \href{link}{https://github.com/LLaVA-VL/LLaVA-NeXT}.

\begin{table}[ht]
\centering
\begin{subtable}[t]{0.48\linewidth}
\centering
\resizebox{\linewidth}{!}{
\sisetup{table-auto-round}
\begin{tabular}{ c c }
\toprule
\textbf{Hyperparameter} & \textbf{Value} \\
\midrule
Training Epochs & 1 \\
Batch Size & 512 \\
Learning Rate (LR) & 1e-3 \\
LR Schedule Type & Cosine\\
Warm-up Ratio & 0.03 \\
Weight Decay & 0.0 \\
\bottomrule
\end{tabular}
}
\caption{Image alignment stage}
\end{subtable}
\hfill
\begin{subtable}[t]{0.48\linewidth}
\centering
\resizebox{\linewidth}{!}{
\sisetup{table-auto-round}
\begin{tabular}{ c c }
\toprule
\textbf{Hyperparameter} & \textbf{Value} \\
\midrule
Training Epochs & 1 \\
Batch Size & 512 \\
Learning Rate (LR) & 1e-5 \\
LR Schedule Type & Cosine\\
Warm-up Ratio & 0.03 \\
Weight Decay & 0.0 \\
\bottomrule
\end{tabular}
}
\caption{High-quality knowledge learning stage}
\end{subtable}
\hfill
\begin{subtable}[t]{0.48\linewidth}
\centering
\resizebox{\linewidth}{!}{
\sisetup{table-auto-round}
\begin{tabular}{ c c }
\toprule
\textbf{Hyperparameter} & \textbf{Value} \\
\midrule
Training Epochs & 1 \\
Batch Size & 512 \\
Learning Rate (LR) & 1e-5 \\
LR Schedule Type & Cosine\\
Warm-up Ratio & 0.03 \\
Weight Decay & 0.0 \\
\bottomrule
\end{tabular}
}
\caption{Visual instruction tuning stage: Single-image training}
\end{subtable}
\hfill
\begin{subtable}[t]{0.48\linewidth}
\centering
\resizebox{\linewidth}{!}{
\sisetup{table-auto-round}
\begin{tabular}{ c c }
\toprule
\textbf{Hyperparameter} & \textbf{Value} \\
\midrule
Training Epochs & 1 \\
Batch Size & 512 \\
Learning Rate (LR) & 1e-5 \\
LR Schedule Type & Cosine\\
Warm-up Ratio & 0.03 \\
Weight Decay & 0.0 \\
\bottomrule
\end{tabular}
}
\caption{Visual instruction tuning stage: OneVision training}
\end{subtable}
\caption{Details of hyperparameters used in LLaVA-OV training}
\label{tab:hyperparameters_ov}
\end{table}

\subsection{Evaluation Results}
\label{sec:app_draft_evaluation}

\cref{tab:app_eval} presents the evaluation results of our primary draft model, LLaVA-1.5 68M, on the OCRBench~\citep{liu2024ocrbenchhiddenmysteryocr} and TextCaps~\citep{sidorov2020textcapsdatasetimagecaptioning} datasets. We assess the output quality of the draft model with and without image inputs and compare the results with those of the target model, LLaVA-1.5 7B. 
In terms of output quality, the draft model with image inputs consistently outperforms the one without, illustrating that the injection of multimodal information benefits the custom-trained draft model.

\cref{fig:ocr} presents qualitative samples from the OCRBench dataset. Both LLaVA-1.5 7B and 68M models provided accurate responses, whereas the text-only LLaVA-1.5 68M model failed to answer correctly due to its lack of image-processing capabilities.
\begin{table}[h]
\centering
\sisetup{table-auto-round}
\resizebox{\linewidth}{!}{%
\begin{tabular}{l|rrr}
\toprule
& OCRBench & \multicolumn{2}{c}{TextCaps}   \\
\cmidrule(lr){1-1}\cmidrule(lr){2-4}
 Model & Accuracy & METEOR & ROUGE \\
 \midrule
LLaVA-1.5 7B  & 0.207 & 0.249 & 0.480 \\
LLaVA-1.5  68M   & 0.048 & 0.133 & 0.254\\
LLaVA-1.5  68M  (text-only)  & 0.014 & 0.064& 0.132\\
\bottomrule
\end{tabular}
}
\caption{
Evaluation results for the off-the-shelf target model and the custom-trained draft model on MLLM tasks. Since the draft model is trained to perceive multimodality, the injection of multimodal information benefits the custom-trained draft model.
}
\label{tab:app_eval}
\end{table}

\tikzset{
  pics/instruction/.style args={#1;#2}{
     code={
        \node[draw, rectangle, rounded corners, 
      fill=yellow!20, text width=5cm, minimum height = 3.5cm,
      align=justify, inner sep=10pt,anchor=south west]
     {};
        \node[rectangle, rounded corners, text width=5cm, 
      align=justify, inner sep=10pt,anchor=south west] at (0,0)
     {#1};
        \node[anchor=south west] at (7pt,1.15) {\includegraphics[width=5cm]{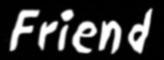}};
     }
  }
}

\tikzset{
  pics/response/.style args={#1}{
     code={
        \node[draw, rectangle, rounded corners, 
      fill=yellow!20, text width=5cm, minimum height = 3.5cm,
      align=justify, inner sep=10pt,anchor=south west]
     {#1};   
     }
  }
}

\begin{figure*}[!htb]
\centering
    \begin{subfigure}[b]{0.24\linewidth}
        \centering
        \resizebox{\linewidth}{!}{%
        \begin{tikzpicture}
        \tikzstyle{every node}=[]
        \draw (0,0) pic{instruction={What is written in the image?;ss}};
        \end{tikzpicture}
        }
        \caption{Instruction}
    \end{subfigure}
    \begin{subfigure}[b]{0.24\linewidth}
        \centering
        \resizebox{\linewidth}{!}{%
        \begin{tikzpicture}
        \tikzstyle{every node}=[]
        \draw (0,0) pic{response={The image has the word ``friend'' written on it.}};
        \end{tikzpicture}
        }
        \caption{LLaVA-1.5 7B}
    \end{subfigure}
    \begin{subfigure}[b]{0.24\linewidth}
        \centering
        \resizebox{\linewidth}{!}{%
        \begin{tikzpicture}
        \tikzstyle{every node}=[]
        \draw (0,0) pic{response={The word ``friend'' is written in the image.}};
        \end{tikzpicture}
        }
        \caption{LLaVA-1.5 68M}
    \end{subfigure}
    \begin{subfigure}[b]{0.24\linewidth}
        \centering
        \resizebox{\linewidth}{!}{%
        \begin{tikzpicture}
        \tikzstyle{every node}=[]
        \draw (0,0) pic{response={The image is a type of text that is written in the image.}};
        \end{tikzpicture}
        }
        \caption{LLaVA-1.5 68M (T)}
    \end{subfigure}
    \caption{Qualitative evaluation samples from the OCRBench dataset by LLaVA-1.5 7B and 68M. Both the target (b) and the draft (c) models recognize the text ``friend'' written on the image by multimodal reasoning whereas the text-only model (d) fails, as expected.}
    \label{fig:ocr}
\end{figure*}
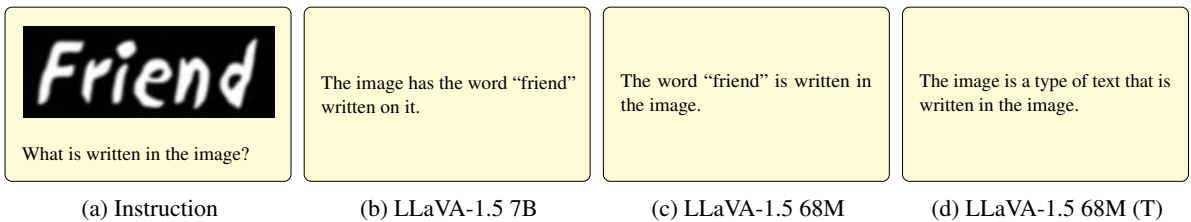

\section{Prompts for Each Dataset and Drafting}
\label{app:system_prompts}
In this section, we describe the formats of prompts used for inference on each dataset, including system prompts and how to organize prompts with text and image inputs (\cref{sec:app_prompt_datasets}). We then provide details on replacing image tokens in text-only and caption drafting (\cref{sec:app_prompt_drafting}). 

\subsection{Prompt Formats for Each Dataset}
\label{sec:app_prompt_datasets}

We use the following prompt formats for their respective tasks. 
Based on the template for chat (\textit{USER:} and \textit{ASSISTANT:}), each system prompt is prepended with the start token \textit{$<$s$>$}.
The \textit{$<$image$>$} token is used to represent image data within a prompt. \textit{[QUESTION]} and \textit{[CAPTION]} are placeholders denoting information unique to each sample of a dataset.

\paragraph{LLaVA-Bench (In-the-Wild)}


\textit{$<$s$>$ USER: $<$image$>$ For the following question, provide a detailed explanation of your reasoning leading to the answer. [QUESTION] ASSISTANT:}


\paragraph{DocVQA}

\textit{$<$s$>$ USER: $<$image$>$ For the following question, provide a detailed explanation of your reasoning leading to the answer. [QUESTION] ASSISTANT:}


\paragraph{POPE}

\textit{$<$s$>$ USER: $<$image$>$ For the following question, provide a detailed explanation of your reasoning leading to the answer. [QUESTION] ASSISTANT:}


\paragraph{MMVet}

\textit{$<$s$>$ USER: $<$image$>$ For the following question, provide a detailed explanation of your reasoning leading to the answer. [QUESTION] ASSISTANT:}


\paragraph{IEdit}

\textit{$<$s$>$ USER: Please provide instructions for editing the source image to match the target image. Source Image: $<$image$>$ Target Image: $<$image$>$ Instruction: ASSISTANT:}

\paragraph{MagicBrush}

\textit{$<$s$>$ USER: Please provide instructions for editing the source image to match the target image. Source Image: $<$image$>$ Target Image: $<$image$>$ Instruction: ASSISTANT:}



\paragraph{Spot The Difference}

\textit{$<$s$>$ USER: Explain the disparities between the first and second image. $<$image$>$ $<$image$>$ Difference: ASSISTANT:}
%

\paragraph{PororoSV}

\textit{$<$s$>$ USER: Given the progression of the story with the first few images, can you write a fitting end considering the last image? $<$image$>$ Caption \#1: [CAPTION] $<$image$>$ Caption \#2: [CAPTION]. $<$image$>$ Caption \#3: [CAPTION] $<$image$>$ Caption \#4: [CAPTION] $<$image$>$ Caption \#5: ASSISTANT:}


\paragraph{VIST}

\textit{$<$s$>$ USER: With the narratives paired with the initial images, how would you conclude the story using the last picture? $<$image$>$ Caption \#1: [CAPTION] $<$image$>$ Caption \#2: [CAPTION]. $<$image$>$ Caption \#3: [CAPTION] $<$image$>$ Caption \#4: [CAPTION] $<$image$>$ Caption \#5: ASSISTANT:}


\subsection{Replacing Image tokens in Draftings}
\label{sec:app_prompt_drafting}
For text-only drafting, the \textit{$<$image$>$} token is replaced by the escape character ``\textbackslash n''. We experimented with several replacement methods: (1) tokenizing the string ``$<$image$>$''  into three tokens, and (2) retaining the special token $<$image$>$ without replacing it with an image embedding. 
 Method (2) resulted in very poor block efficiency, but method (1) showed comparable block efficiency. Our replacement approach is simple because it ensures that the prompt length remains consistent before and after replacement.

 For caption drafting, the \textit{$<$image$>$} token is replaced by a generated caption with a prefix. Specifically, after the lightweight captioning model generates a caption based on the image inputs in the sample, we prepend the string ``image: '' to the caption and replace the \textit{$<$image$>$} token.

\section{Details of Each Dataset}

In this section, we describe each of the curated datasets in benchmark (\cref{sec:app_benchmark_datasets}) and OOD (\cref{sec:app_ood_datasets}) datasets and provide links to them for convenience and reproducibility.

\label{app:benchmark}







\subsection{Benchmark Datasets}
\label{sec:app_benchmark_datasets}

\paragraph{LLaVA-Bench (In-the-Wild) \citep{liu2023llava}}
A dataset for comparing the performance of vision-language models against state-of-the-art proprietary models. Each prompt is provided with an image, a caption, and a reference answer from text-only GPT-4. Prompt styles include question answering, image description, and complex reasoning. In total, the dataset contains 60 unique prompts and 24 unique images.

\url{https://huggingface.co/datasets/lmms-lab/llava-bench-in-the-wild}

\paragraph{DocVQA \citep{mathew2021docvqa}}
A dataset designed for visual question answering on document images, comprising 50,000 questions over 12,000+ diverse document images.

\url{https://huggingface.co/datasets/lmms-lab/LMMs-Eval-Lite}

\paragraph{POPE \citep{li2023evaluating}}
A multimodal question answering dataset that asks binary (yes or no) questions about whether certain objects are present in an image. The subset used for evaluation in our work contains 100 pairs of images and questions.

\url{https://huggingface.co/datasets/lmms-lab/LMMs-Eval-Lite}

\paragraph{MMVet \citep{yu2023mm}}
A benchmark designed to evaluate large multimodal models on complex tasks, focusing on the integration of six core vision-language capabilities: recognition, OCR, knowledge, language generation, spatial awareness, and math.

\url{https://huggingface.co/datasets/lmms-lab/MMVet}

\paragraph{Spot the Difference \citep{jhamtani2018learning}}
A dataset of crowd-sourced descriptions of differences between a pair of images. The subset used for evaluation in our work contains 100 annotated image pairs collected using individual frames of security-footage data. 
This dataset is available under the Creative Commons Attribution 4.0 International license.

\url{https://huggingface.co/datasets/lmms-lab/LLaVA-NeXT-Interleave-Bench}

\paragraph{IEdit \citep{tan2019expressing}}
A dataset to train models to describe the relationship between images via editing instructions. The subset used for evaluation in our work contains 100 image pairs of a source image and a target image, accompanied by instructions on how to transform the source image into the target.
This dataset is available under the Creative Commons Attribution 4.0 International license.

\url{https://huggingface.co/datasets/lmms-lab/LLaVA-NeXT-Interleave-Bench}

\paragraph{MagicBrush \citep{zhang2024magicbrush}}
A dataset for text-guided image editing containing manually annotated editing instructions to transform one real image into another. The subset used for evaluation in our work contains 100 triplets of a source image, a target image, and editing instructions.
This dataset is available under the Creative Commons Attribution 4.0 International license.

\url{https://huggingface.co/datasets/lmms-lab/LLaVA-NeXT-Interleave-Bench}

\subsection{OOD Datasets}
\label{sec:app_ood_datasets}

\paragraph{Pororo-SV \citep{li2019storygan}}
A dataset of stories each created by pairing 5 consecutive frames from the animated series \textit{Pororo} with a text description. The subset used for evaluation in our work contains 100 stories.
This dataset is available under the Creative Commons Attribution 4.0 International license.

\url{https://huggingface.co/datasets/lmms-lab/LLaVA-NeXT-Interleave-Bench}

\paragraph{VIST \citep{huang2016visual}}
A dataset of sequential images paired with three types of descriptions ranging from isolated factual descriptions to causal, narrative interpretations. The subset used for evaluation in our work contains 100 sequences of 3 images.
This dataset is available under the Creative Commons Attribution 4.0 International license.

\url{https://huggingface.co/datasets/lmms-lab/LLaVA-NeXT-Interleave-Bench}
\section{Time Analysis of LVLM Inference Stages}
\label{app:inference-time}
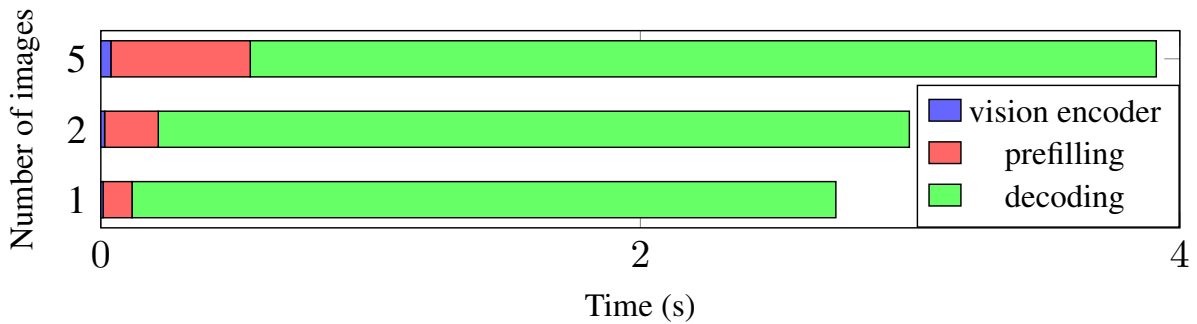
\begin{figure*}[t]
    \centering
            \resizebox{1\linewidth}{!}{%
\begin{tikzpicture}
\begin{axis}[
    width=12cm,
    height=3.5cm,
    xbar stacked,    
    bar width=10pt,
    xlabel={\small{Time (s)}},
    ylabel={\small{Number of images}},
    yticklabels={1,2,5},    
    ytick=data,             
    xtick={0,2,4},         
    xmin=0,                
    xmax=4,                
    legend style={
        at={(1,0)},
        anchor=south east,
    },
    symbolic y coords={1,2,5},    
    enlarge y limits=0.2,         
]
\addplot[fill=blue!60] coordinates {
    (0.008683240414,1) (0.01402390718,2) (0.03731709719,5)    
};
\addplot[fill=red!60] coordinates {
    (0.1070554304,1) (0.1989724088,2) (0.5167041755,5)        
};
\addplot[fill=green!60] coordinates {
    (2.6091325117,1) (2.7842734818,2) (3.3586801395,5)           
};
\legend{\small\vision{}, \small\prefill{}, \small\decode{}}
\end{axis}
\end{tikzpicture}

}
\captionsetup{skip=2pt}
\caption{Inference time analysis for the LLaVA-1.5 7B model. Although the time for vision encoder and prefilling  increases with the number of images, the decoding stage still dominates.} 
\label{fig:timespan}
\end{figure*}

In this section, we present the time analysis of the inference stages of LVLMs.
Specifically, \cref{fig:timespan} analyzes how the number of input images affects the LVLM inference time, we select ChartQA \citep{masry2022chartqa}, Spot the Difference \citep{jhamtani2018learning}, and PororoSV \citep{li2019storygan} datasets representing 1, 2, and 5 images with corresponding visual context lengths of 0.6k, 1.2k, and 3k, respectively. We visualize the generation time by component in \cref{fig:timespan} with 100 generated tokens for analysis, with actual average decoding lengths of 92, 117, and 88, respectively. The execution time of the \textit{\vision{}} and \textit{\prefill{}} stages increases in proportion with the number of input images, as each image is converted into several hundred context tokens. In contrast, the \textit{\decode{}} stage shows little difference in execution time across varying numbers of input images, while dominating the total generation time. Hence, although reducing the number of visual tokens \citep{shang2024llava, chen2024image, lin2024boosting} would significantly improve the efficiency of \textit{\vision{}} and \textit{\prefill{}} stages, it would have only marginal impact on the dominant \textit{\decode{}} stage.
\begin{table*}[t]
\centering
\sisetup{table-auto-round}
\renewcommand{\arraystretch}{1.0}
\resizebox{1.0\textwidth}{!}{%
\begin{tabular}{lrl|ccccccc|c||cc}
\toprule
\multicolumn{3}{c|}{Draft Model} & \multicolumn{8}{c||}{\titlecap{\standard{} datasets} (First Turn)} & \multicolumn{2}{c}{\titlecap{\ood{} datasets}} \\
\cmidrule(lr){1-3}\cmidrule(lr){4-11}\cmidrule(lr){12-13}
Type & Size & Method & LLaVA-W & DocVQA & POPE & MMVet & IEdit & MB & Spot &Avg.& PSV & VIST \\
\midrule
\multirow{8}{*}{\shortstack[l]{LLaVA-1.5}} & \multirow{8}{*}{68M} & M &2.28 & 2.15 & 2.56 & 2.21 & 2.19 & 1.96 & 2.34 &2.24& 1.19& 1.16\\
 &  & T & 2.19 & 2.08 & 2.31 & 2.16 & 2.23 & 2.34 & 2.27 &2.23&2.05 &2.05\\
 &&C &2.22&	2.15&	2.50&	2.17&	2.29&	2.36&	2.31&	2.29&	2.08&	2.10\\
 &&P &2.22&	2.08&	2.42&	2.17&	2.22&	2.27&	2.23&2.23	&2.07&	2.09\\
  &  & \textbf{MT} & 2.25 & 2.15 & 2.47 & 2.21 & 2.31 & 2.37 & 2.40 &2.31& 1.94	& 1.91\\
 &  & \textbf{MT*} & 2.26 & 2.16 & 2.52 & 2.21 & 2.29 & 2.39 & 2.36 &2.31& 2.02 & 2.04\\
&&\textbf{MTCP}&2.27&2.15&	2.49&	2.22 &2.32	&2.39&	2.40&2.32&	2.10&	2.13\\
&&\textbf{MTCP*}&2.29&2.17	&2.51&	2.22	&2.30	&2.40	&2.35	&2.32&2.11	&2.14\\
\midrule
\multicolumn{3}{c|}{Draft Model} & \multicolumn{8}{c||}{\titlecap{\standard{} datasets} (Second Turn)} & \multicolumn{2}{c}{\titlecap{NLP Datasets}} \\
\cmidrule(lr){1-3}\cmidrule(lr){4-11}\cmidrule(lr){12-13}
Type & Size & Method & LLaVA-W & DocVQA & POPE & MMVet & IEdit & MB & Spot &Avg.& NQ & GSM8K \\
\midrule
\multirow{8}{*}{\shortstack[l]{LLaVA-1.5}} & \multirow{8}{*}{68M} & M & 2.1 & 1.96 & 2.78 & 2.18 & 1.61 & 1.53 & 1.83 &2.00& 1.98 & 2.25 \\
 &  & T & 2.32 & 2.23 & 2.91 & 2.56 & 1.87 & 2.01 & 2.08  &2.28& 2.03 & 2.30 \\
  &&C &2.28 & 2.29 & 2.94 & 2.58 & 1.87 & 2.01 & 2.10 &2.30& 2.03 &2.29\\
 &&P&2.28&2.09&2.95&2.49& 1.86 & 2.00 & 2.08&2.25&2.01&2.26\\
  &  & \textbf{MT} & 2.29 & 2.24 & 2.93 & 2.54 & 1.81 & 1.91 & 2.02  &2.25& 2.02 & 2.28 \\
 &  & \textbf{MT*}  & 2.29 & 2.23 & 2.93 & 2.56 & 1.85 & 1.99 & 2.05  &2.27& 2.03 & 2.29 \\
&&\textbf{MTCP} & 2.32 & 2.29 & 3.03 & 2.59 & 1.86 & 2.00 & 2.08&2.31&2.02&2.27\\
&&\textbf{MTCP*}& 2.34 & 2.31 & 3.04 & 2.59 & 1.86 & 2.01 & 2.09 & 2.32 & 2.02 & 2.28\\
\bottomrule
\end{tabular}
}
\captionsetup{skip=5pt}
\caption{
Full experimental results with LLaVA-1.5 68M (draft model) and LLaVA-1.5 7B (target model)
}
\label{tab:rebt1}
\end{table*}

\begin{table*}[t]
\centering
\sisetup{table-auto-round}
\renewcommand{\arraystretch}{1.0}
\resizebox{1.0\textwidth}{!}{%
\begin{tabular}{lrl|ccccccc|c||cc}
\toprule
\multicolumn{3}{c|}{Draft Model} & \multicolumn{8}{c||}{\titlecap{\standard{} datasets} (First Turn)} & \multicolumn{2}{c}{\titlecap{\ood{} datasets}} \\
\cmidrule(lr){1-3}\cmidrule(lr){4-11}\cmidrule(lr){12-13}
Type & Size & Method & LLaVA-W & DocVQA & POPE & MMVet & IEdit & MB & Spot &Avg.& PSV & VIST \\
\midrule
\multirow{8}{*}{\shortstack[l]{LLaVA-1.5}} & \multirow{8}{*}{68M} & M & 2.01 & 1.96 & 2.58 & 2.14 & 1.89 & 1.77 & 2.05 & 2.06 & 1.19 & 1.15 \\
 &  & T & 1.97 & 1.92 & 2.27 & 2.11 & 1.85 & 2.12 & 2.03 & 2.04 & 1.79 & 1.90 \\
  &  & C & 2.00 & 1.97 & 2.48 & 2.14 & 1.92 & 2.16 & 2.04 & 2.10 & 1.78 & 1.93\\
  &  & P & 2.00 & 1.90 & 2.41 & 2.11 & 1.87 & 2.05 & 1.97 & 2.04 & 1.78 & 1.92\\
  &  & \textbf{MT} & 2.02 & 1.97 & 2.47 & 2.14 & 1.94 & 2.14 & 2.08 & 2.11 & 1.75 & 1.80\\
 &  & \textbf{MT*} & 2.02 & 1.97 & 2.50 & 2.14 & 1.94 & 2.15 & 2.07 & 2.11 & 1.79 & 1.89\\
   &  & \textbf{MTCP} & 2.02 & 1.99 & 2.50 & 2.15 & 1.94 & 2.18 & 2.09 & 2.12 & 1.81 & 1.93\\
&  & \textbf{MTCP*} & 2.03 & 1.99 & 2.52 & 2.15 & 1.94 & 2.19 & 2.08 & 2.13 & 1.82 & 1.95\\
\midrule
\multicolumn{3}{c|}{Draft Model} & \multicolumn{8}{c||}{\titlecap{\standard{} datasets} (Second Turn)} & \multicolumn{2}{c}{\titlecap{NLP Datasets}} \\
\cmidrule(lr){1-3}\cmidrule(lr){4-11}\cmidrule(lr){12-13}
Type & Size & Method & LLaVA-W & DocVQA & POPE & MMVet & IEdit & MB & Spot &Avg.& NQ & GSM8K \\
\midrule
\multirow{8}{*}{\shortstack[l]{LLaVA-1.5}} & \multirow{8}{*}{68M} & M & 1.88 & 1.88 & 2.78 & 2.08 & 1.52 & 1.43 & 1.67 & 1.89 & 1.94 & 2.12\\
 &  & T & 2.05 & 2.09 & 2.84 & 2.45 & 1.76 & 1.83 & 1.93 & 2.14 & 1.99 & 2.17\\
    &  & C & 2.07 & 2.12 & 2.92 & 2.48 & 1.75 & 1.84 & 1.94 & 2.16 & 1.99 & 2.17\\
   &  & P & 2.04 & 1.97 & 2.89 & 2.39 & 1.75 & 1.82 & 1.93 & 2.11 & 1.96 & 2.13\\
  &  & \textbf{MT} & 2.03 & 2.08 & 2.88 & 2.41 & 1.69 & 1.75 & 1.86 & 2.10 & 1.97 & 2.15\\
 &  & \textbf{MT*}  & 2.04 & 2.10 & 2.88 & 2.42 & 1.71 & 1.78 & 1.87 & 2.11 & 1.97 & 2.15\\
    &  & \textbf{MTCP} & 2.06 & 2.13 & 2.95 & 2.46 & 1.74 & 1.82 & 1.94 & 2.16 & 1.98 & 2.16\\
&  & \textbf{MTCP*} & 2.07 & 2.14 & 2.97 & 2.47 & 1.75 & 1.83 & 1.95 & 2.17 & 1.98 & 2.16\\
\bottomrule
\end{tabular}
}
\captionsetup{skip=5pt}
\caption{
Full experimental results with LLaVA-1.5 68M (draft model) and LLaVA-NeXT 7B (target model)
}
\label{tab:rebt2}
\end{table*}

\begin{table*}[t]
\centering
\sisetup{table-auto-round}
\renewcommand{\arraystretch}{1.0}
\resizebox{1.0\textwidth}{!}{%
\begin{tabular}{lrl|ccccccc|c||cc}
\toprule
\multicolumn{3}{c|}{Draft Model} & \multicolumn{8}{c||}{\titlecap{\standard{} datasets} (First Turn)} & \multicolumn{2}{c}{\titlecap{\ood{} datasets}} \\
\cmidrule(lr){1-3}\cmidrule(lr){4-11}\cmidrule(lr){12-13}
Type & Size & Method & LLaVA-W & DocVQA & POPE & MMVet & IEdit & MB & Spot &Avg.& PSV & VIST \\
\midrule
\multirow{8}{*}{\shortstack[l]{LLaVA-1.5}} & \multirow{8}{*}{160M} & M & 2.29 & 2.29 & 3.06 & 2.44 & 2.17 & 2.04 & 2.26 & 2.36 & 1.23 & 1.24\\
 &  & T & 2.25 & 2.19 & 2.56 & 2.36 & 2.20 & 2.41 & 2.24 & 2.32 & 1.96 & 2.08\\
  &  & C & 2.26 & 2.26 & 2.87 & 2.42 & 2.26 & 2.46 & 2.29 & 2.40 & 2.05 & 2.19\\
  &  & P & 2.29 & 2.23 & 2.94 & 2.42 & 2.27 & 2.49 & 2.32 & 2.42 & 2.08 & 2.20\\
  &  & \textbf{MT} & 2.31 & 2.30 & 2.95 & 2.44 & 2.24 & 2.46 & 2.32 & 2.43 & 1.89 & 1.97 \\
 &  & \textbf{MT*} & 2.26 & 2.31 & 3.01 & 2.44 & 2.19 & 2.49 & 2.31 & 2.43 & 1.95 & 2.07\\
   &  & \textbf{MTCP} & 2.31 & 2.30 & 3.02 & 2.44 & 2.30 & 2.54 & 2.34 & 2.46 & 2.06 & 2.20\\
 &  & \textbf{MTCP*} & 2.31 & 2.31 & 3.06 & 2.45 & 2.31 & 2.54 & 2.33 & 2.47 & 2.07 & 2.22\\
\midrule
\multicolumn{3}{c|}{Draft Model} & \multicolumn{8}{c||}{\titlecap{\standard{} datasets} (Second Turn)} & \multicolumn{2}{c}{\titlecap{NLP Datasets}} \\
\cmidrule(lr){1-3}\cmidrule(lr){4-11}\cmidrule(lr){12-13}
Type & Size & Method & LLaVA-W & DocVQA & POPE & MMVet & IEdit & MB & Spot &Avg.& NQ & GSM8K \\
\midrule
\multirow{8}{*}{\shortstack[l]{LLaVA-1.5}} & \multirow{8}{*}{160M} & M & 2.15 & 2.14 & 3.17 & 2.45 & 1.74 & 1.6 & 1.94 & 2.17 & 2.28 & 2.48 \\
 &  & T & 2.37 & 2.31 & 3.04 & 2.72 & 2.11 & 2.14 & 2.33 & 2.43 & 2.3 & 2.55\\
    &  & C & 2.40 & 2.38 & 3.11 & 2.75 & 2.12 & 2.16 & 2.33 & 2.46 & 2.29 & 2.56\\
   &  & P & 2.39 & 2.35 & 3.25 & 2.76 & 2.08 & 2.13 & 2.33 & 2.47 & 2.29 & 2.51\\
  &  & \textbf{MT} & 2.37 & 2.37 & 3.21 & 2.77 & 2.04 & 2.06 & 2.23 & 2.44 & 2.29 & 2.54\\
 &  & \textbf{MT*} & 2.34 & 2.36 & 3.21 & 2.75 & 2.11 & 2.18 & 2.27 & 2.46 & 2.30 & 2.53 \\
   &  & \textbf{MTCP} & 2.42 & 2.38 & 3.22 & 2.78 & 2.08 & 2.13 & 2.30 & 2.47 & 2.30 & 2.54\\
 &  & \textbf{MTCP*} & 2.42 & 2.39 & 3.22 & 2.79 & 2.09 & 2.14 & 2.31 & 2.48 & 2.30 & 2.54\\
\bottomrule
\end{tabular}
}
\captionsetup{skip=5pt}
\caption{
Full experimental results with LLaVA-1.5 160M (draft model) and LLaVA-NeXT 7B (target model)
}
\label{tab:rebt3}
\end{table*}

\begin{table*}[t]
\centering
\sisetup{table-auto-round}
\renewcommand{\arraystretch}{1.0}
\resizebox{1.0\textwidth}{!}{%
\begin{tabular}{lrl|ccccccc|c||cc}
\toprule
\multicolumn{3}{c|}{Draft Model} & \multicolumn{8}{c||}{\titlecap{\standard{} datasets} (First Turn)} & \multicolumn{2}{c}{\titlecap{\ood{} datasets}} \\
\cmidrule(lr){1-3}\cmidrule(lr){4-11}\cmidrule(lr){12-13}
Type & Size & Method & LLaVA-W & DocVQA & POPE & MMVet & IEdit & MB & Spot &Avg.& PSV & VIST \\
\midrule
\multirow{8}{*}{\shortstack[l]{LLaVA-1.5}} & \multirow{8}{*}{160M} & M & 1.96 & 1.93 & 2.22 & 2.10 & 1.91 & 1.98 & 1.98 & 2.01 & 1.65 & 1.73\\
 &  & T & 1.99 & 1.91 & 2.02 & 2.10 & 1.86 & 2.24 & 1.99 & 2.02 & 1.81 & 1.86 \\
  &  & C & 2.00 & 1.96 & 2.17 & 2.13 & 1.92 & 2.20 & 1.98 & 2.05 & 1.79 & 1.84\\
  &  & P & 1.99 & 1.91 & 2.13 & 2.10 & 1.96 & 2.19 & 2.02 & 2.04 & 1.87 & 1.91\\
  &  & \textbf{MT} & 2.01 & 1.95 & 2.21 & 2.14 & 1.93 & 2.22 & 2.02 & 2.07 & 1.84 & 1.89\\
 &  & \textbf{MT*} & 2.01 & 1.96 & 2.23 & 2.13 & 1.96 & 2.27 & 1.99 & 2.08 & 1.83 & 1.88\\
   &  & \textbf{MTCP} & 2.02 & 1.97 & 2.28 & 2.14 & 1.97 & 2.26 & 2.08 & 2.10 & 1.88 & 1.93 \\
 &  & \textbf{MTCP*} & 2.02 & 1.97 & 2.29 & 2.14 & 1.97 & 2.27 & 2.07 & 2.10 & 1.88 & 1.93 \\
\midrule
\multicolumn{3}{c|}{Draft Model} & \multicolumn{8}{c||}{\titlecap{\standard{} datasets} (Second Turn)} & \multicolumn{2}{c}{\titlecap{NLP Datasets}} \\
\cmidrule(lr){1-3}\cmidrule(lr){4-11}\cmidrule(lr){12-13}
Type & Size & Method & LLaVA-W & DocVQA & POPE & MMVet & IEdit & MB & Spot &Avg.& NQ & GSM8K \\
\midrule
\multirow{8}{*}{\shortstack[l]{LLaVA-1.5}} & \multirow{8}{*}{160M} & M & 1.97 & 1.96 & 2.37 & 2.11 & 1.63 & 1.64 & 1.77 & 1.92 & 1.87 & 2.16 \\
 &  & T & 2.06 & 2.08 & 2.57 & 2.39 & 1.79 & 1.86 & 1.96 & 2.10 & 1.97 & 2.23 \\
    &  & C & 2.07 & 2.12 & 2.67 & 2.43 & 1.79 & 1.86 & 1.94 & 2.13 & 1.96 & 2.23 \\
   &  & P & 2.04 & 2.09 & 2.62 & 2.44 & 1.78 & 1.85 & 1.93 & 2.11 & 1.94 & 2.23 \\
  &  & \textbf{MT} & 2.07 & 2.10 & 2.60 & 2.41 & 1.75 & 1.81 & 1.90 & 2.09 & 1.95 & 2.23 \\
 &  & \textbf{MT*} & 2.07 & 2.10 & 2.60 & 2.42 & 1.76 & 1.83 & 1.91 & 2.10 & 1.95 & 2.23 \\
   &  & \textbf{MTCP} & 2.09 & 2.14 & 2.71 & 2.45 & 1.78 & 1.86 & 1.93 & 2.14 & 1.96 & 2.24 \\
 &  & \textbf{MTCP*} & 2.09 & 2.14 & 2.69 & 2.46 & 1.79 & 1.86 & 1.93 & 2.14 & 1.96 & 2.24\\
\bottomrule
\end{tabular}
}
\captionsetup{skip=5pt}
\caption{
Full experimental results with LLaVA-OV 68M (draft model) and LLaVA-NeXT 7B (target model)
}
\label{tab:rebt4}
\end{table*}

\begin{table*}[t]
\centering
\sisetup{table-auto-round}
\renewcommand{\arraystretch}{1.0}
\resizebox{1.0\textwidth}{!}{%
\begin{tabular}{lrl|ccccccc|c||cc}
\toprule
\multicolumn{3}{c|}{Draft Model} & \multicolumn{8}{c||}{\titlecap{\standard{} datasets} (First Turn)} & \multicolumn{2}{c}{\titlecap{\ood{} datasets}} \\
\cmidrule(lr){1-3}\cmidrule(lr){4-11}\cmidrule(lr){12-13}
Type & Size & Method & LLaVA-W & DocVQA & POPE & MMVet & IEdit & MB & Spot &Avg.& PSV & VIST \\
\midrule
\multirow{8}{*}{\shortstack[l]{LLaVA-1.5}} & \multirow{8}{*}{68M} & M & 2.01 & 1.97 & 2.44 & 2.04 & 1.87 & 1.75 & 1.95 & 2.00 & 1.18 & 1.15 \\
 &  & T & 1.98 & 1.93 & 2.23 & 2.02 & 1.86 & 2.07 & 1.94 & 2.00 & 1.73 & 1.89\\
  &  & C & 1.98 & 1.96 & 2.40 & 2.05 & 1.89 & 2.09 & 1.92 & 2.04 & 1.74 & 1.90 \\
  &  & P & 1.97 & 1.91 & 2.35 & 2.02 & 1.87 & 1.98 & 1.87 & 2.00 & 1.74 & 1.90 \\
  &  & \textbf{MT} & 2.00 & 1.97 & 2.36 & 2.06 & 1.91 & 2.07 & 1.99 & 2.05 & 1.69 & 1.79 \\
 &  & \textbf{MT*} & 2.02 & 1.94 & 2.39 & 2.06 & 1.89 & 2.07 & 1.97 & 2.05 & 1.73 & 1.86\\
   &  & \textbf{MTCP} & 2.00 & 1.99 & 2.39 & 2.06 & 1.93 & 2.11 & 2.00 & 2.07 & 1.75 & 1.91\\
 &  & \textbf{MTCP*} & 2.01 & 1.99 & 2.40 & 2.06 & 1.93 & 2.11 & 1.99 & 2.07 & 1.75 & 1.93 \\
\midrule
\multicolumn{3}{c|}{Draft Model} & \multicolumn{8}{c||}{\titlecap{\standard{} datasets} (Second Turn)} & \multicolumn{2}{c}{\titlecap{NLP Datasets}} \\
\cmidrule(lr){1-3}\cmidrule(lr){4-11}\cmidrule(lr){12-13}
Type & Size & Method & LLaVA-W & DocVQA & POPE & MMVet & IEdit & MB & Spot &Avg.& NQ & GSM8K \\
\midrule
\multirow{8}{*}{\shortstack[l]{LLaVA-1.5}} & \multirow{8}{*}{68M} & M & 1.85 & 1.82 & 2.73 & 2.02 & 1.52 & 1.43 & 1.72 & 1.87 & 1.97 & 2.18 \\
 &  & T & 2.00 & 2.03 & 2.70 & 2.31 & 1.77 & 1.81 & 1.98 & 2.09 & 2.01 & 2.22 \\
    &  & C & 2.03 & 2.08 & 2.77 & 2.35 & 1.77 & 1.81 & 2.00 & 2.12 & 2.01 & 2.23 \\
   &  & P & 2.01 & 1.91 & 2.75 & 2.27 & 1.76 & 1.79 & 1.98 & 2.07 & 1.99 & 2.19 \\
  &  & \textbf{MT} & 2.02 & 2.02 & 2.74 & 2.29 & 1.69 & 1.73 & 1.91 & 2.06 & 1.99 & 2.20 \\
 &  & \textbf{MT*} & 2.01 & 2.01 & 2.79 & 2.30 & 1.74 & 1.78 & 1.95 & 2.08 & 2.00 & 2.22 \\
   &  & \textbf{MTCP} & 2.04 & 2.07 & 2.82 & 2.32 & 1.74 & 1.79 & 1.97 & 2.11 & 2.00 & 2.21 \\
 &  & \textbf{MTCP*} & 2.04 & 2.07 & 2.82 & 2.32 & 1.75 & 1.79 & 1.98 & 2.11 & 2.00 & 2.22 \\
\bottomrule
\end{tabular}
}
\captionsetup{skip=5pt}
\caption{
Full experimental results with LLaVA-1.5 68M (draft model) and LLaVA-NeXT 13B (target model)
}
\label{tab:rebt5}
\end{table*}

\end{document}